%% file: main_corl.tex
\documentclass{article}

\usepackage{subcaption}

\usepackage[final]{corl_2020} 

\usepackage{booktabs}
\usepackage{graphicx}
\usepackage[toc,page]{appendix}

\def\shownotes{1}  
\ifnum\shownotes=1
\newcommand{\authnote}[2]{{$\ll$\textsf{\footnotesize #1 notes: #2}$\gg$}}
\else
\newcommand{\authnote}[2]{}
\fi

\usepackage{footnote}
\newcommand\blfootnote[1]{%
  \begingroup
  \renewcommand\thefootnote{}\footnote{#1}%
  \addtocounter{footnote}{-1}%
  \endgroup
}

\title{Learning to Compose Hierarchical Object-Centric Controllers for Robotic Manipulation}

\author{
    Mohit Sharma$^{\dag 1}$\\
    Robotics Institute\\
    Carnegie Mellon University \\
    \And
    Jacky Liang$^{\dag 1}$\\
    Robotics Institute \\
    Carnegie Mellon University \\
    \And
    Jialiang Zhao$^1$ \\  
    Robotics Institute \\
    Carnegie Mellon University \\
    \And
    Alex LaGrassa$^1$ \\
    Robotics Institute \\
    Carnegie Mellon University \\
    \And
    Oliver Kroemer$^1$ \\
    Robotics Institute \\
    Carnegie Mellon University \\
}
\usepackage{amsmath}
\usepackage{amssymb}
\usepackage{bbm}
\usepackage{algorithm,algorithmic}
\usepackage{url}
\usepackage{color}
\usepackage{graphicx}
\usepackage{caption}
\usepackage{hyperref}

\newcommand{\todo}[1]{}
\renewcommand{\todo}[1]{{\color{red}Todo: {#1}}}

\definecolor{alizarin}{rgb}{0.82, 0.1, 0.26}
\definecolor{table_color_train}{rgb}{0.682, 0.612, 0.271}
\definecolor{table_color_test}{rgb}{0.376, 0.451, 0.694}

\begin{document}
\maketitle
{\centering
$^1$\texttt{\{mohitsharma, jackyliang, alanjz, alagrass, okroemer\}@cmu.edu}\par
}
\vspace{-3mm}
\blfootnote{$^{\dag}$ Equal Contribution}

\input{includes/0_abstract}
\input{includes/1_intro}
\input{includes/2_rw}

\input{includes/3_method}
\input{includes/4_exps_setup}
\input{includes/5_exps_results}
\input{includes/6_conclusion}

\input{includes/7_ack}


\bibliography{citations}  

\clearpage
\input{includes/8_appendix.tex}

\end{document}

%% file: includes/0_abstract.tex
\begin{abstract}
Manipulation tasks can often be decomposed into multiple subtasks performed in parallel, \emph{e.g.}, sliding an object to a goal pose while maintaining contact with a table. 
Individual subtasks can be achieved by task-axis controllers defined relative to the objects being manipulated, and a set of object-centric controllers can be combined in an hierarchy.
In prior works, such combinations are defined manually or learned from demonstrations.
By contrast, we propose using reinforcement learning to dynamically compose hierarchical object-centric controllers for manipulation tasks.
Experiments in both simulation and real world show how the proposed approach leads to improved sample efficiency, zero-shot generalization to novel test environments, and simulation-to-reality transfer without fine-tuning.

\end{abstract}

%% file: includes/1_intro.tex
\section{Introduction}
\label{sec:intro}

Manipulation tasks are inherently object-centric and often require a robot to perform multiple subtasks in parallel, such as pressing on a sponge while wiping across a surface, balancing a saucer while serving tea, or maintaining alignment of a screwdriver while unscrewing a screw. 
The individual subtasks need to be performed in parallel to accomplish the overall task.  
As the above examples illustrate, subtasks usually correspond to goals and constraints associated to objects in the robot’s environment. 
Thus, manipulation skills are often defined as 3D motions, which are implemented as simple position or force controllers, of the end effector in object-centric coordinate frames.

\begin{figure}[!t]
    \centering
    \includegraphics[width=\linewidth]{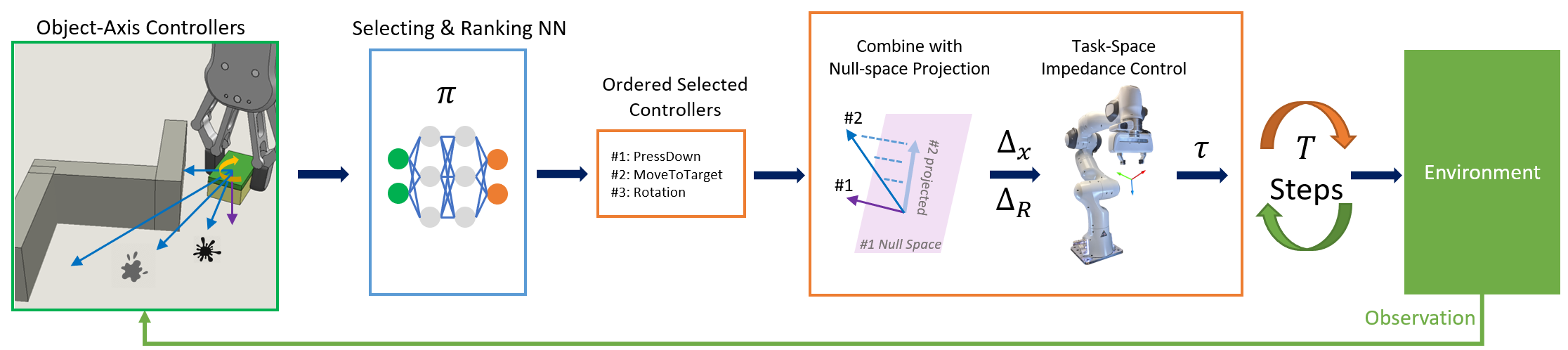}
    \caption{
    \footnotesize{
    Controller Selection and Composition Pipeline. 
    Given current observations and list of low-level controllers, an RL policy chooses an ordered list of controllers to use. 
    These controllers are composed via nullspace projection, where the controls of lower-priority controllers are projected onto the nullspace of higher-priority ones.
    The combined control signals are used to actuate a robot via task-space impedance control.
    The controller combination runs for $T$ time steps before the RL policy is queried again.
    }}
    \label{fig:pipeline}
    \vspace{-15pt}
\end{figure}

One drawback of such an approach is that it results in monolithic controllers for each task, \emph{i.e.} controllers which act specifically with respect to some fixed coordinate frame.
In addition, for many tasks it is not always necessary to control all axes of a given object-centric coordinate frame. 
For instance, for the wiping task in Figure~\ref{fig:pipeline}, the sponge needs to use the table surface normal to make contact with the surface, while it is free to move with respect to any other object (wall, corners, dirt) on the surface.
Based on this insight, we adopt a modular approach by defining task-axis controllers for each potential subtask.
Importantly, the controllers are associated with object-centric axes, such as the normal of a surface or the direction from the end-effector to an object.

We focus on learning an hierarchy of such object-centric task-axis controllers, or \textbf{object-axis controllers} (Figure~\ref{fig:pipeline}).
This hierarchy is especially important since many tasks require performing multiple subtasks in parallel. 
Previous works use pre-defined sets of task frames attached to objects or the robot, and they often learn a fixed task-frame hierarchy from human demonstrations.
Instead, we use Reinforcement Learning (RL) to learn a policy that outputs an ordered list of controllers, which are then composed to be executed on the robot. 
To ensure different object-axis controllers do not interfere with each other, we compose controllers via nullspace projections~\cite{abu2015adaptation}, where the control signals of lower-priority controllers are projected onto the nullspace of higher priority ones.

In addition to modularity, our approach provides several other benefits.
First, the object-axis controllers are not task specific, so they can be reused across multiple tasks. 
Second, composing controllers across multiple different objects makes the learned policies invariant to certain object properties \emph{e.g.}, a controller that reaches toward an object is invariant to object size. 
Such invariances are useful for generalizing learned policies beyond the set of objects the policies are trained on.
Finally, the use of a structured action space introduces meaningful inductive biases by ensuring robot actions are performed both in relation and with respect to objects in the scene.
We successfully evaluated our approach on four different manipulation tasks, including two 2D tasks of fitting and pushing a block and two real robot tasks of screwing and door-opening. 
Experiments show that the proposed approach leads to improved sample efficiency, zero-shot generalization to novel environment configurations, and simulation-to-reality transfer without further fine-tuning.
See videos and supplementary materials at \url{https://sites.google.com/view/compositional-object-control/}.

%% file: includes/2_rw.tex
\section{Related Works}
\label{sec:rw}


\textbf{Task Frames:} 
Our use of task-axes is related to the notion of 6D task frames \cite{ballard1984task, ballard1986task, mason1981compliance}.
One of the first works to formalize task frames is~\cite{mason1981compliance}.
There, the authors referred to different task-axes as compliant or non-compliant based on the type of desired motion along each axis. 
The authors of~\cite{raibert1981hybrid} proposed hybrid force-position control, which selects different axes of the constraint frame for either position or force control.
Simultaneously, the authors of~\cite{ballard1984task, ballard1986task} proposed task frames to define robotic manipulation primitives; they noted that the geometric level of task frames can serve as a good middle ground between symbolic actions and the motor control input.
Since then, task frames in the form of task spaces have been used extensively in robotics \cite{sciavicco2012modelling}.
Prior works treat task-frames as fixed coordinate frames which are either attached to objects of interest or generated from constraints in the environment. 
By contrast, our approach is more modular and dynamic, as it enables an RL policy to combine task-axes across different objects and dynamically synthesize task-frames.

\textbf{Task Frame Selection:} 
Although the use of task frames and spaces is widespread in robotics \cite{muhlig2009automatic, berenson2011task, king2016rearrangement, kober2015learning, ureche2015task, migimatsu2020object, manschitz2020learning}, only a few works have explored using learning to select which task frames are appropriate for the given task \cite{muhlig2009automatic, kober2015learning, ureche2015task, peternel2017method, conkey2019learning}.
However, most of these works use imitation learning \emph{i.e.}, they learn task frame selection from human demonstrations \cite{muhlig2009automatic, kober2015learning, ureche2015task, peternel2017method}.
The criterion for task-frame selection is typically manually defined using properties such as inter-trial variance or convergence behavior of demonstrations.
In our work, we set task-axes selection as the action space for an RL agent, so we do not require demonstrations.
Moreover, the RL agent chooses a hierarchy of task-axis controllers, which are composed together for execution. 

\textbf{Hierarchical Controllers:} 
Combining multiple task-axis controllers is related to works in hierarchical control.
Hierarchical control is often used in robots with redundant degrees of freedom or bi-manual robot setups where multiple tasks or objectives can be executed in parallel \cite{khatib1987unified, nakamura1987task, dietrich2015overview, karami2018hierarchical}. 
To combine different controllers, these works project the control signals of lower-priority controllers onto the nullspace of higher-priority controllers.
However, most of these works assume a fixed priority order for the tasks/objectives being considered, while some recent works~\cite{karami2018hierarchical} learn the priorities from human demonstrations. 
Similar to these works, our approach also uses nullspace projections to combine multiple task-axis controllers together. 
However, instead of using a fixed priority order, our method learns to prioritize controllers by directly interacting with the environment.

\textbf{Reinforcement Learning:} 
Finally, our approach is related to works on structured action spaces for reinforcement learning (RL) for contact-rich manipulation tasks. 
Recent works have studied how the choice of action spaces affect robot learning performance \cite{martin2019variable, bogdanovic2019learning, beltran2020learning}. 
However, these methods focus only on the final controller output, \emph{i.e.}, comparing fixed with variable impedance control~\cite{martin2019variable, bogdanovic2019learning} or with hybrid-force position control~\cite{beltran2020learning} in joint and task-spaces. 
Our work provides additional structure to the action space via composing hierarchical object-centric controllers.

\textbf{Hierarchical RL:}
Composing task-axes controllers for performing tasks is also related with hierarchical RL (HRL) \cite{sutton1999between, comanici2010optimal, konidaris2009efficient}.
HRL uses the notion of options, which are temporally-extended actions, and learns to combine them to accomplish a given task. There has been a large body of work which aims to extract the underlying options \cite{stolle2002learning, bacon2017option, csimcsek2005identifying}, 
using techniques such as bottleneck states \cite{stolle2002learning}, policy sketches \cite{shiarlis2018taco}, or expert demonstrations 
\cite{krishnan2017ddco, sharma2018directedinfo, daniel2016probabilistic}.
Similarly, there have also been works that use predefined option policies and compose them to learn a ``meta-policy" \citep{liaw2017composing}.
However, these option policies are defined specifically with respect to the underlying task, 
and hence it is not clear how reusable these policies are.
By contrast, our proposed task-axes controllers are reusable across multiple different manipulation tasks. 
This is desirable for efficient learning of new manipulation tasks\cite{morrow_khosla_1997}.
Additionally, task-axes controllers are different than options since they can be composed both hierarchically and temporally.

%% file: includes/3_method.tex
\section{Learning Hierarchical Compositions of Object-Centric Controllers}
\label{sec:method}

\begin{figure}[!t]
    \centering
    \includegraphics[width=\linewidth]{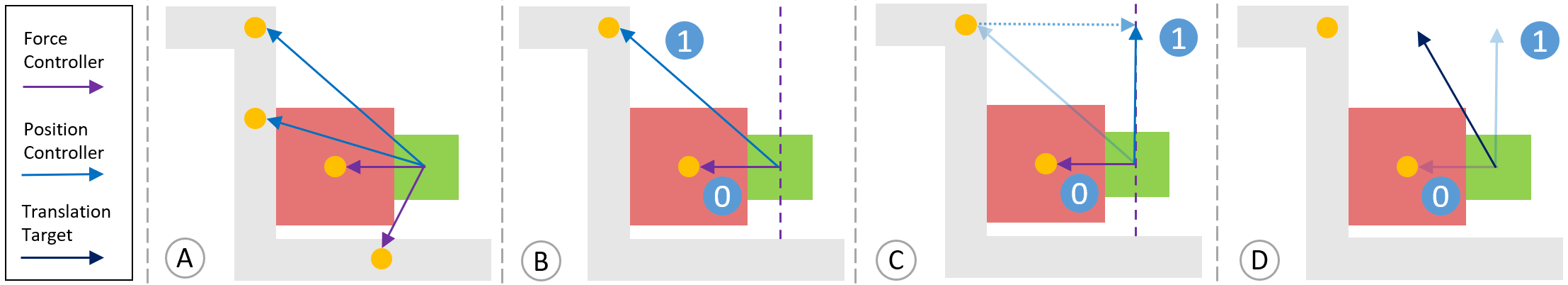}
    \caption{
    \footnotesize{
    Force-Position Controller Composition.
    Here, the agent controls the green block to push the red block up along the vertical gray wall.
    A) The agent is given $4$ controllers to choose from, each corresponding to points of interests in the scene.
    B) The agent chooses $2$ controllers, with the force controller into the red block at the higher priority ($0$), and position controller toward the wall corner at the lower priority ($1$).
    C) The error of the lower-priority position controller is projected onto the null space of the higher-priority force controller (purple dashed line).
    D) The projected errors are combined to form the desired position target.
    }
    }
    \label{fig:ctrlr_comp_fp}
    \vspace{-15pt}
\end{figure}



We propose training an RL policy to perform manipulation tasks by using a structured action space consisting of hierarchical compositions of object-centric controllers.
Each object in the scene is associated with a fixed set of task-axes, positioned either at object centers or other object key points.
For each axis, we define a set of controllers that perform force, position, and rotation controls.
This gives a set of pre-defined object-centric task-axis controllers, or object-axis controllers, which define our structured action space.
With this action space, instead of directly commanding the end-effector, the RL policy selects multiple object-axis controllers in a prioritized order, which are composed together using null-space projections. 
Figure~\ref{fig:pipeline} shows an overview of the overall proposed approach.

In the next subsections, we first define the different types of object-centric low-level controllers we use, including how their object-centric axes are defined. 
We then discuss how to combine different object-axis controllers together using null-space projections. 
Finally, we discuss different RL approaches for learning the high-level policy that selects multiple controllers.

\subsection{Controller Types}
In this work, we use three different types of controllers: position, force, and rotation.
These controllers are object-centric, \emph{i.e.} their control targets and axes correspond to objects in the scene.
For example, position controllers could be attractors that lead the end-effector (EE) close to an object of interest, force controllers could be applying forces perpendicular to object surfaces, and rotation controllers could be aligning an axis of the EE with an axis of the object.
Currently, these controllers are manually specified (see details in Section~\ref{sec:exps}), but they could also be autonomously inferred from visual observations of objects in the environment.
Figure~\ref{fig:ctrlr_comp_fp} illustrates force and position controllers and their composition, and Figure~\ref{fig:ctrlr_comp_rot} shows the rotation controllers.

Let $x_c \in \mathbb{R}^3$, $R_c \in SO(3)$, and $f_c \in\mathbb{R}^3$ respectively denote the current end-effector position, orientation, and forces expressed in the robot's base frame.

\textbf{Position and Force Controllers:}
The position controller consists of a target position $x_d$ and an axis $u$ along which the controller will move the robot's end-effector toward the target.
$u$ can be a fixed direction, like the normal direction of a surface, or it can be adapted with respect to $x_c$: $u = \frac{x_d - x_c}{\|x_d - x_c\|_2}$.
Let $\mathcal{P}(u) = u u^\top$ be the projection matrix for the given axis. 
Then, the translation error a position controller produces is defined as $\delta_x(x_d, u, x_c) = \mathcal{P}(u) (x_d - x_c)$.
The force controller is similar to the position controller, \emph{i.e.} given a force target $f_d$ and an axes-direction $u$, the force error the controller produces is $\delta_f(f_d, u, f_c) = \mathcal{P}(u) (f_d - f_c)$.

\textbf{Rotation Controller:}
The rotation controller attempts to align one axis $R_cu$ of $R_c$ with a target axis $r_d$, where $u$ is a unit vector that performs axis-selection.
For example,to align the X-axis of the end-effector frame to align with $r_d$, then $u = [1, 0, 0]^\top$.
The rotation controller produces a delta rotation target in the end-effector frame, which we compute via the angle-axis representation: $\delta_R(r_d, u, R_c) = \cos^{-1}( (R_c u)^\top r_d) ((R_c u) \times r_d)$

\textbf{Null Controllers:}
The high-level policy also has the option to choose a null controller, which would give $0$ errors for both $\delta_x$ and $\delta_R$.
While other controllers can be chosen at most $1$ time, the null controllers can be chosen multiple times, giving the high-level policy more flexibility.

\subsection{Controller Composition}

\textbf{Force-Position Composition:}
The RL policy selects at most $3$ force and position controllers to compose.
Only $3$ of force and position controllers can execute concurrently, because there are only $3$ position dimensions.
The RL policy outputs a priority order for these controllers.
Let the indices $[0, 1, 2]$ denote the $3$ controllers in decreasing priority, so $0$ is the highest, and $2$ the lowest.
The final position target is computed by projecting the lower-priority targets onto the nullspaces of the higher-priority controllers, then summing them.
Let $\mathcal{N}(U) = I - U^\dagger U$ be a nullspace projection matrix with respect to rows of $U$, where $\dagger$ denotes the pseudoinverse.
Let $K_x$ be the position controller gain and $K_f$ the force gain:
\begin{align}
    \Delta_x^0 &= K_x\delta_x(x_d^0, u^0, x_c)\\
    \Delta_x^1 &= K_x\mathcal{N}([u^0]) \delta_x(x_d^1, u^1, x_c) \\
    \Delta_x^2 &= K_x\mathcal{N}([u^0, u^1]) \delta_x(x_d^2, u^2, x_c)\\
    \Delta_x &= \sum_{i=0}^2 \Delta_x^i
\end{align}
where $[\hdots]$ represents a concatenation operator, \emph{i.e.} concatenation of vectors into a matrix, \emph{e.g.}, $[u^0, u^1] \in \mathbb{R}^{2\times3}$.
Although the above expressions are written with all $3$ controllers as position controllers, in our implementation we combine multiple position and force controllers together.
If force controllers are used, for the corresponding controller, swap $\delta_x$ with $\delta_f$, $x_d$ with $f_d$,  $x_c$ with $f_c$, and $K_x$ with $K_f$.
Figure~\ref{fig:ctrlr_comp_fp} illustrates the force-position controller composition.

\begin{figure}[!t]
    \centering
    \includegraphics[width=\linewidth]{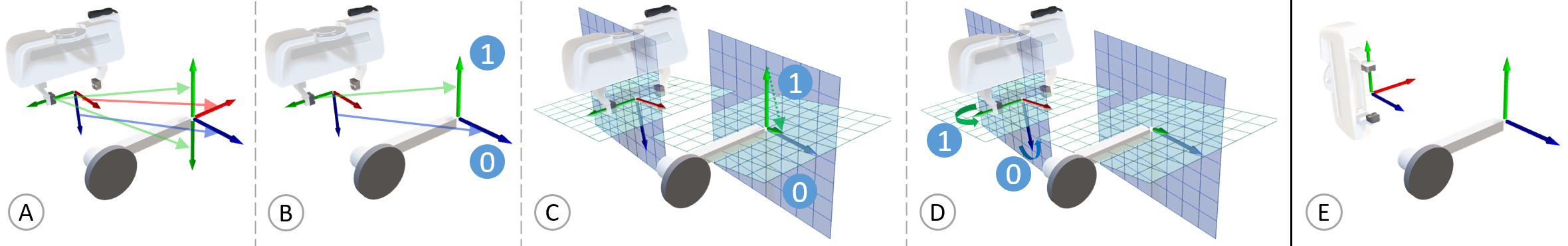}
    \caption{
    \footnotesize{
    Rotation Controller Composition. 
    Here, the agent rotates the Franka robot's gripper from the initial pose (A) to the final pose (E), so the gripper aligns with a door handle.
    A) The agent is given $4$ rotation controllers to choose from, aligning various axes of the gripper with different target axes of the handle.
    B) Two controllers are chosen with the higher-priority labeled as ($0$) and the lower-priority as ($1$).
    C) Both the current and target axes of the lower-priority controller (green arrows) are projected down to the null-space (green planes) of the current axis of the higher-priority controller (gripper's blue axis).
    D) The desired rotation target is formed by combining the higher-priority rotation in the blue plane with the projected lower-priority rotation in the green plane. Note that the lower-priority rotation does not interfere with the higher-priority rotation.}
    }
    \label{fig:ctrlr_comp_rot}
    \vspace{-15pt}
\end{figure}

\textbf{Rotation Composition:}
The RL policy selects at most two rotation controllers to compose.
This is because when the highest priority controller fixes one axis of a rotation frame, there is only one degree of freedom left, which is a rotation in the 2D nullspace of the fixed axis.
Similar to force-position controller compositions, we project the errors of lower-priority controllers onto the nullspace of higher-priority controllers:
\begin{align}
    \Delta_R^0 &= K_R \delta_R(r_d^0, u^0, R_c)\\
    \Delta_R^1 &= K_R \delta_R(\mathcal{N}([R_c u^0]) r_d^1, u^1, \mathcal{N}([R_c u^0]) R_c)\\
    \Delta_R &= \Delta_R^1 \circ \Delta_R^0
\end{align}
where $\circ$ denotes composing rotations, and $K_R$ denotes a rotation error gain.
This procedure ensures the higher-priority rotation controller always reaches its goal, and the trajectory of that axis is not affected by the lower-priority controller (see Figure~\ref{fig:ctrlr_comp_rot} for an illustration).

\textbf{Controlling the Robot:}
We use task-space impedance control to convert translation and rotation targets to configuration-space targets via Jacobian transpose, and we actuate the robot via joint torques.
We first concatenate the translation target $\Delta_x$ with the axis-angle representation of $\Delta_R$ to form the final 6D delta end-effector target $\Delta$.
Then, the robot joint-torque commands are computed as $\tau = J^\top (K_S \Delta + K_D \dot{\Delta})$,
where $K_S$ and $K_D$ are diagonal stiffness and damping matrices, and $J$ is the analytic Jacobian.
Terms for compensating gravity and Coriolis forces are omitted for brevity.
In practice, we cap the magnitude of $\Delta$ to limit maximum control effort, and we add an integral term to the force controllers for better convergence. 
Once a set of controllers are selected, their combination runs for $T$ timesteps before the RL policy is queried again for a new set of controllers.

\subsection{RL with Object-Axis Controllers}
\label{sec:rl_task_axis_controllers}

We use RL to learn a policy that composes object-axis controllers to perform the underlying task. 
The policy outputs an ordered list of controllers, which are composed together to output the final control signal to move the robot.
The combination of controllers is run for a fixed $T$ timesteps, before the RL policy is queried again.
Note that the controllers do not have to converge before the RL policy switches to the next combination.
We next discuss multiple ways in which the RL policy can output the ordered list of controllers.

\textbf{Discrete Combinatorial Actions:} 
Let $N$ be the total number of available controllers, and $N_c$ be the number of controllers that can be executed simultaneously. 
One simple way to output an ordered list of $N_c$ controllers is to use a discrete action space, where the policy selects an action from all available controller permutations. 
Such an action space grows combinatorially ($\mathcal{O}(N^{N_c})$), and is not scalable for environments with a large number of controllers.

\textbf{Continuous Priority Scores:}  
A continuous space alternative is to allow the policy to output a priority score in $[0, 1]$ for all controllers. 
These priority scores are then used to order the controllers, where the $N_c$ controllers with highest priorities are executed at each step. 
Although the dimension of this action space grows linearly with the number of controllers, it can often lead to sub-optimal performance since the agent now needs to explore a much larger action space than before.

\textbf{Expanded-MDP:} 
To avoid the sub-optimal performance of the above methods, we propose an expanded-MDP formulation that still uses a discrete action space while avoiding combinatorial expansion.
Here, we expand each environment-execution step of the MDP into $N_c$ intermediate controller-selection steps, with the original environment-execution step occurring after the $N_c$'th intermediate step.
At each intermediate step, the policy selects one controller from the $N$ choices. 
Once $N_c$ controllers are selected, the robot takes an actual environment step.
The reward function is modified such that $0$ rewards are given for the controller-selection steps before the $N_c$'th step.
Similar MDP transformations have been suggested previously to solve continuous action MDPs using discrete action space RL algorithms ~\cite{pazis2011reinforcement, metz2017discrete}.


To use the Expanded-MDP formulation, at each controller-selection step the policy needs to know its previous controller selections.
One approach is representing each controller with 1-hot encoding and appending the 1-hot encodings of previously selected controllers to the observations.
This expands the observation space by $N\times(N_c-1)$ dimensions, and we refer to this representation as \textbf{multi-1-hot}.
However, in many cases it might not be necessary to know the order of the previous controllers being selected, \emph{i.e.}, it is sufficient to know which controllers have been selected previously but not their order. 
So, for the second representation, we merge the one-hot encodings of multiple previous controllers into one binary vector.
This only increases the observation space by $N$ dimensions, and can lead to faster learning.
We refer to this representation as \textbf{single-1-hot}

%% file: includes/4_exps_setup.tex
\section{Experiment Tasks and Setup}
\label{sec:exps}

With our experiments we aim to evaluate 1) How useful are the proposed object-axis controllers for task learning, 2) How important is controller composition for task learning, and 3) How well does our proposed approach generalize to the different test configurations.

\begin{figure}[!t]
    \centering
    \includegraphics[width=0.754\linewidth]{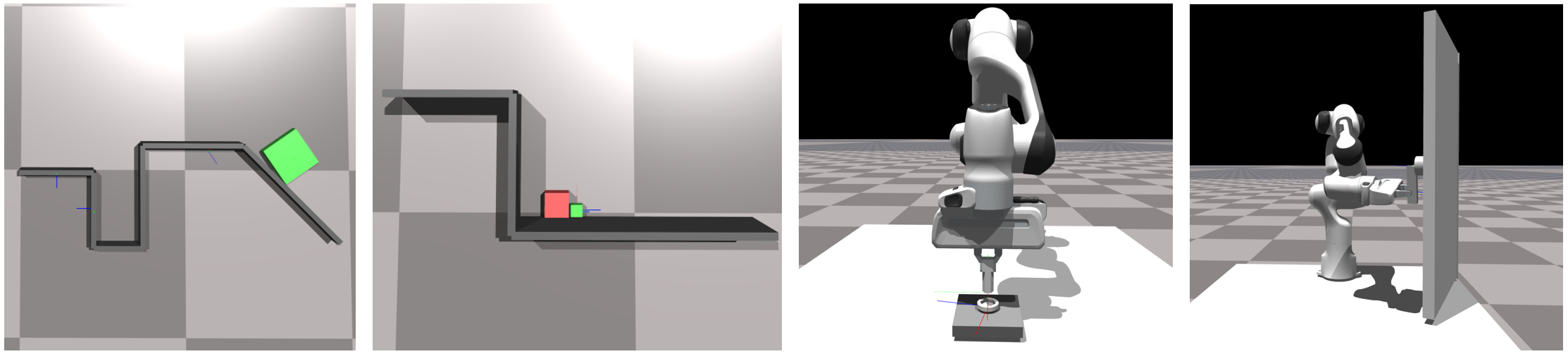}
    \includegraphics[width=0.108\linewidth]{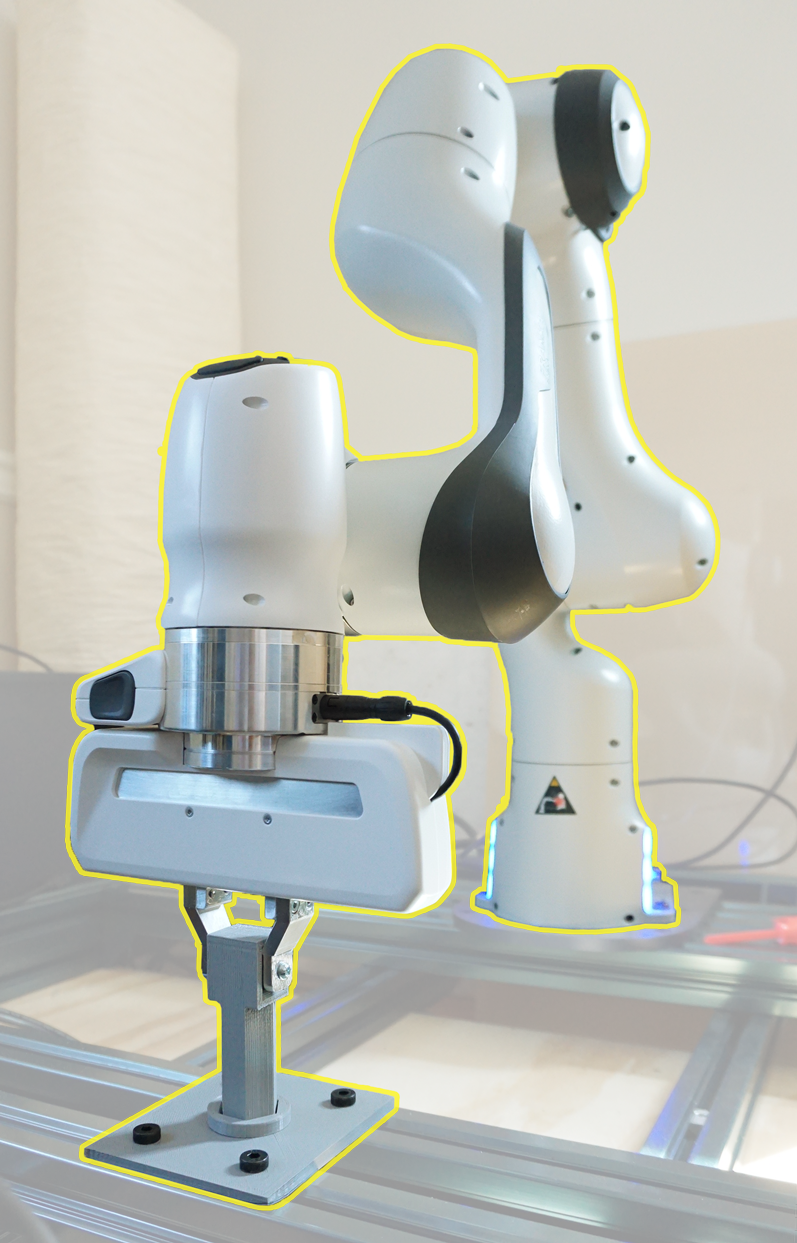}
    \includegraphics[width=0.118\linewidth]{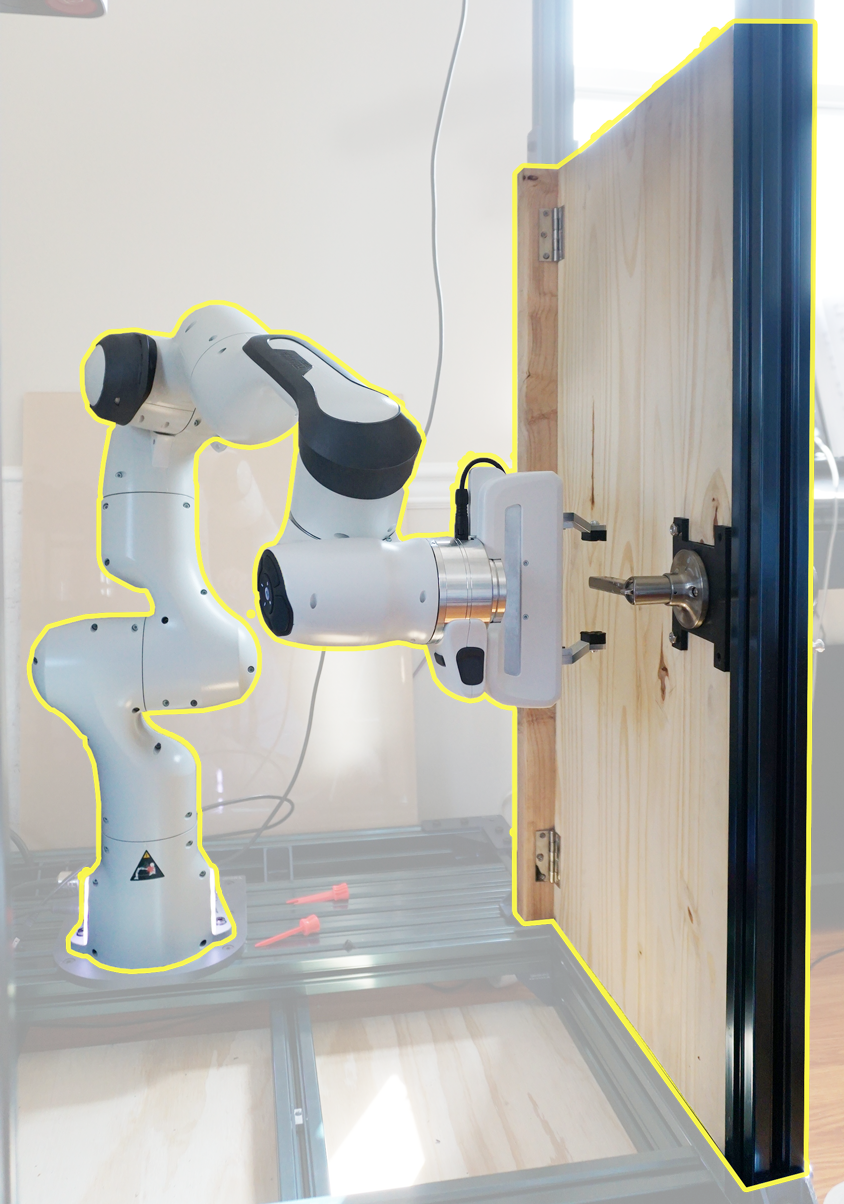}
    \caption{\footnotesize{Experiment Tasks. From left to right: Block Fit, Block Push, Franka Hex-Screw, Franka Door-Opening tasks implemented in simulation, and Franka tasks in the real world.}}
    \label{fig:tasks}
    \vspace{-10pt}
\end{figure}

Figure~\ref{fig:tasks} visualizes the tasks used to evaluate our approach. 
There are two 2D tasks, Block Fit and Block Push, and two real robot tasks, screwing hex-screws and opening doors with the 7 DoF Franka Emika Panda arm.
We compare both learning performance of the proposed approach against baselines, as well as their ability to generalize to novel environment configurations.
To study generalization, we train policies on a small set of training environment configurations and test them on a novel test set.
Training over multiple environments is important to avoid overfitting.
Details of each task, including controller specifications, task variations, observation and action spaces, and the reward functions can be found in the Appendix.

\begin{figure}[t]
    \centering
    \includegraphics[width=0.8\linewidth]{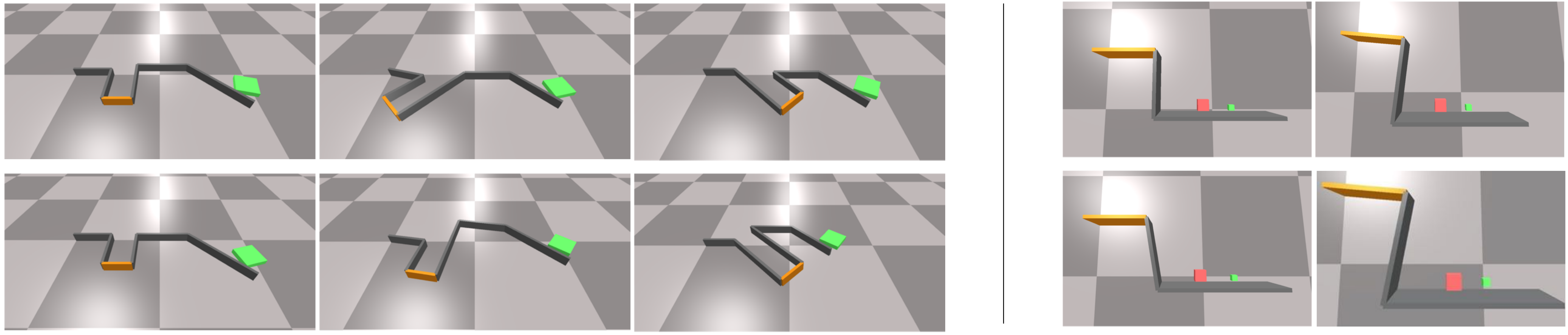}
    \caption{\footnotesize{Example environment configurations for Block Push (left) and Block Fit (right) environments. Top row shows \emph{some} examples of train configurations, and the bottom row shows \emph{some} examples of test configurations. The orange wall shows the goal wall to reach.}}
    \label{fig:variations_block_envs}
    \vspace{-12pt}
\end{figure}

\textbf{Block Fit:}
In this task, a 2D block robot needs to navigate to a 2D goal pose in the scene.
There are multiple walls or obstacles in the scene, so the robot cannot directly proceed towards the goal.
Figure~\ref{fig:variations_block_envs} (Left) shows \emph{some} of the different train and test configurations.
The low-level controllers are wall-centric.
Different environment configurations have different wall lengths and angles between walls.
The training set has $8$ different environment configurations, while the test set has $9$.

\textbf{Block Push:}
In this task, a 2D block robot needs to push another block along a vertical wall over a ledge to a desired goal pose.
Figure~\ref{fig:variations_block_envs} (right) visualizes \emph{some} train and test configurations.
Controllers and environment wall configurations are similar to those of Block Fit.
The environment samples the initial pose of the block robot and the target block.
The training set has $11$ different environment configurations, and the test set has $8$.

\textbf{Franka Hex-Screw:}
In this task, a 7-DoF Franka Panda arm is used to insert a hex-key into a screw, and turn the screw to a desired angle while applying a downward force and maintaining vertical orientation.
The screw will not turn unless a sufficient pre-defined ($20N$) downward force is applied.
Different environment configurations have different wrench and screw sizes. 
The training set uses size scale multipliers of $(0.9, 1.0, 1.3)$, and the test set uses $(0.7, 0.8, 1.1, 1.2, 1.4, 1.5)$.  

\textbf{Franka Door-Opening:}
In this task, the Franka robot needs to open a door by first turning its door handle and then pulling the door beyond an opening threshold.
To avoid trivial policy solutions, the door will not open unless the handle is first turned to a desired angle.
The environment samples the initial relative pose between the EE and the door, and different configurations have different locations of the door handle on the door.
The training and test set contain $4$ and $3$ configurations.

\textbf{Compared Approaches:}
We set $N_c = 3$ across all experiments, which we found to be sufficient.
To evaluate the utility of our proposed object-axis controllers we compare against an RL agent that controls the robot directly via end-effector delta-poses. We call this approach \textbf{EE-Space}.
We also evaluate the need for executing multiple controllers in parallel by comparing against a baseline which only chooses 1 controller at each timestep. We call this \textbf{1-Ctrlr}.
To show the efficacy of our proposed Expanded-MDP formulation we compare against both: discrete combinatorial (\textbf{3-Combo}) and continuous priority scores (\textbf{3-Priority}) action spaces. 
Both these approaches naively combine all possible controller combinations and we show how this can lead to sub-optimal performance.


\textbf{RL Training:}
We use Proximal Policy Optimization (PPO)~\cite{schulman2017proximal} implemented in stable-baselines~\cite{stablebaselines} across all tasks and action space variants.
Given the high variance in policy-gradient RL algorithms, we run all methods with $8$ different seeds (sampled uniformly between $1$ and $100$). 
All tasks are simulated with an NVIDIA Isaac Gym~\footnote{\url{https://developer.nvidia.com/isaac-gym}}, a GPU-accelerated robotics simulator~\cite{liang2018gpu}.

\textbf{Metrics:}
We report the success rates of the learned policies separately for train and test environment configurations.
Performance on the train set indicates whether or not the approach can robustly solve a task, and performance on the test set evaluates generalization abilities.
Test set is split into two subsets, one with small deviations from the train configurations, and another with larger deviations.
We report additional results including more fine-grained analysis for each task in the Appendix.

%% file: includes/5_exps_results.tex
\section{Experiment Results and Discussion}
\label{sec:results}

\begin{figure}[t]
    \centering
    \includegraphics[width=0.96\linewidth]{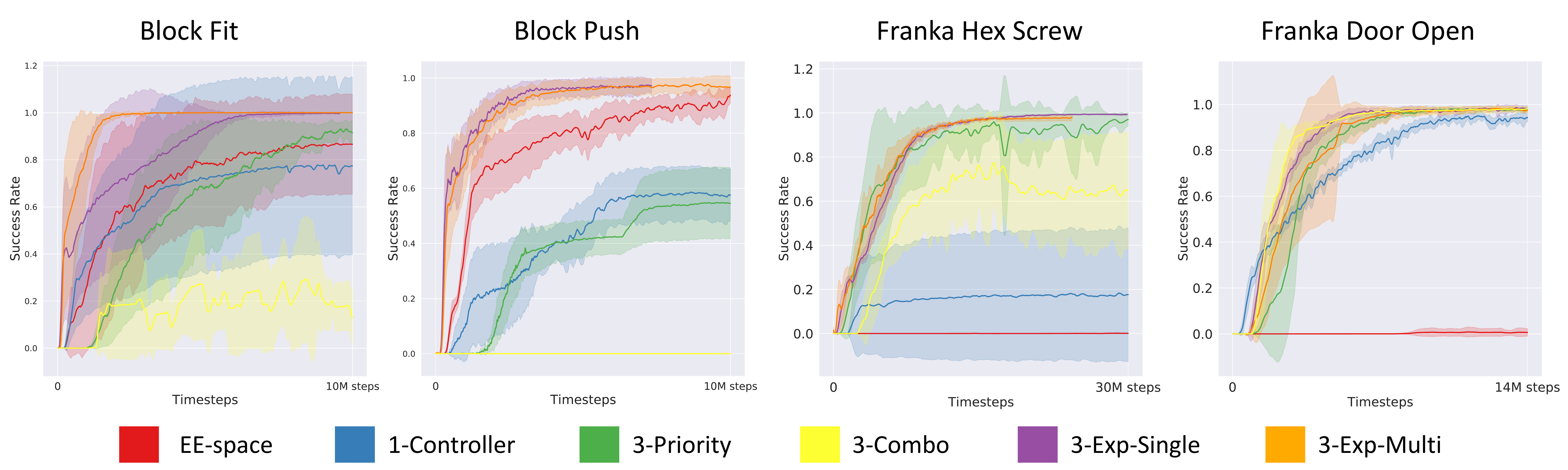}
    \caption{\footnotesize{Success rates for all tasks on training environment configurations.}}
    \label{fig:all_env_results}
    \vspace{-15pt}
\end{figure}

\textbf{Block Tasks:}
Figure~\ref{fig:all_env_results} (left) plots the success ratios averaged over all train environment configurations for Block Fit and Block Push.
The Expanded-MDP methods are able to successfully learn both tasks.
While EE-Space also makes progress on both tasks, it has a lower success rate, and this is due to its inability to robustly solve a few challenging configurations (see Appendix).
Both 1-Ctrlr and 3-Priority perform well on Block Fit but poorly on Block Push.
We attribute this difference to how there is a greater need to use multiple controllers in the right order for Block Push.
For instance, the policy needs to choose a force/position controller that pushes into the wall and then another controller to move up. 
In addition, robustly pushing the block around the edge of the vertical wall also requires multiple controllers.
Although it is feasible to achieve this by quickly switching between controllers, such a strategy is not robust.
1-Ctrlr is unable to use multiple controllers at the same time, and using the high-dimensional priority score action space is challenging.

Table~\ref{tab:block2d_train_test_results} shows success rates for both tasks on two sets of test configurations. 
Both EE-space and Expanded-MDP methods perform well when test configurations have small deviations from train configurations, with EE-Space performing slightly worse.
However, for large deviations, EE-space performs poorly, achieving success ratios of $0.371$ for Block Fit and $0.518$ for Block Push. 
By contrast, Expanded-MDP methods perform much better, achieving $0.974$ for Block Fit and $0.788$ for Block Push, and 3-Priority also outperforms EE-Space for the Block Fit task.
In addition, 1-Ctrlr sees greater performance degradation going from small to large deviations in test configurations.
Together, these results indicate that using a structured action space of multiple object-centric controllers leads to better generalization than using one controller or directly learning in the EE-space.

\begin{table}[]
\centering
\resizebox{\textwidth}{!}{%
\begin{tabular}{@{}lllllllll@{}}
\toprule
Task       & Variation  & EE-Space              & 1-Ctrlr                       & 3-Priority        & 3-Combo             & 3-Exp-Single                   & 3-Exp-Multi  \\ 
\midrule
Block Fit  & Train      & 0.87 \small{(0.213)}  & 0.778 \small{(0.38)}     & 0.936 \small{(0.032)} & 0.294 (0.18)       & 0.998 \small{(0.002)}          & \textbf{1.00 \small{(0.0)}}\\
 & Test-Small & 0.87 \small{(0.10)}   & 0.916 \small{(0.14)} & \textbf{0.99 \small{(0.001)}}  & 0.184 (0.12)   & \textbf{0.99 \small{(0.001)}}  &  \textbf{0.99 \small{(0.01)}}\\
 & Test-Large & 0.371 \small{(0.246)} & 0.396 \small{(0.423)}         &  0.877 \small{(0.141)}  & 0.165 (0.23)       & \textbf{0.974 \small{(0.048)}} & 0.953 \small{(0.087)}   \\
\midrule
Block Push & Train      & 0.966 \small{(0.046)} & 0.594 \small{(0.087)}         & 0.548 \small{(0.129)}     & 0.0 (0.0)        & 0.974 \small{(0.025)}          & \textbf{0.978 \small{(0.022)}}\\
           & Test-Small & 0.912 \small{(0.045)} & 0.577 \small{(0.193)}         & 0.396 \small{(0.041)}     & 0.0 (0.0)      & 0.945 \small{(0.045)}          & \textbf{0.960 \small{(0.030)}} \\
           & Test-Large & 0.518 \small{(0.185)} & 0.152 \small{(0.137)}         & 0.376 \small{(0.032)}     & 0.0 (0.0)     & 0.751 \small{(0.103)}          & \textbf{0.788 \small{(0.132)}} \\ 
\bottomrule \\
\end{tabular}%
}
\caption{\footnotesize{Mean (SD) success rates for Block Fit and Block Push tasks on different environment configurations.}}
\label{tab:block2d_train_test_results}
\vspace{-3mm}
\end{table}

\begin{table}[]
\centering
\resizebox{\textwidth}{!}{%
\begin{tabular}{@{}llllllll@{}}
\toprule
Task      & Variation   & EE-Space                & 1-Ctrlr                & 3-Priority         & 3-Combo           & 3-Exp-Single                   & 3-Exp-Multi                    \\ 
\midrule
Hex-Screw & Train       & 0.002 \small{(0.002)}   & 0.183 \small{(0.303)}  & 0.960 \small{(0.048)} & 0.774 \small{0.194)} & \textbf{0.984 \small{(0.01)}}  & 0.980 \small{(0.016)}           \\
          & Test-Small  & 0.00 (0.00)             & 0.13 (0.072)          & 0.62 (0.045)         & 0.429 \small{(0.430)}          & 0.963 \small{(0.01)}          & \textbf{0.966 \small{(0.015)}}  \\
          & Test-Large  & 0.00 \small{(0.00)}     & 0.026 \small{(0.025)}  & 0.633 \small{(0.081)} & 0.34 (0.057)   & \textbf{0.936 \small{(0.028)}} & \textbf{0.936 \small{(0.035)}} \\
          & Real-World  & n/a                     & 0.0                    & 0.5                 & 0.0     & \textbf{0.9}                            & 0.6                         \\
\midrule
Door-Open & Train       & 0.002 \small{(0.006)}   & 0.947 \small{(0.021)}  & 0.982 \small{(0.007)}  & 0.984 \small{(0.013)} & \textbf{0.987 \small{(0.009)}} & 0.984 \small{(0.015)} \\
          & Test-Small  & 0.066 \small{(0.063)}   & 0.922 \small{(0.043)}  & 0.965 \small{(0.046)} & 0.975 \small{(0.011)}  & \textbf{0.997 \small{(0.006)}} & 0.992 \small{(0.015)} \\
          & Test-Large  & 0.000 \small{(0.001)}   & 0.936 \small{(0.032)}  & 0.983 \small{(0.006)} & 0.985 \small{(0.007)} & \textbf{0.996 \small{(0.005)}} & 0.994 \small{(0.013)}           \\ 
          & Real-World  & n/a                     & 0.0                    & \textbf{1.0}       & 0.9    & \textbf{1.0}                            & \textbf{1.0}                       \\
\bottomrule \\
\end{tabular}%
}
\caption{\footnotesize{
Success rates for Franka Hex-Screw and Open-Door tasks on train and test environment configurations across $8$ seeds. 
Parentheses denote standard deviation.
Real-world results are evaluated over $10$ trials each.
We did not run EE-Space policies in the real world as they were unable to learn the tasks in simulation.
}}
\label{tab:franka_all_env_results}
\vspace{-10mm}
\end{table}

\textbf{Franka Tasks:}
Figure~\ref{fig:all_env_results} (right) shows training results for both Franka Hex-Screw and Door-Open tasks.
The Expanded-MDP methods perform well on both the tasks, while EE-Space does not make progress on either task.
For Hex-Screw, the EE-Space policy is able to reach the screw, but is unable to learn to simultaneously rotate the screw and apply sufficient downward force.
For Door-Open, the EE-Space policy reaches the door handle, but fails to grasp and completely rotate the door handle in a robust manner to open the door.
One reason for these EE-Space failures is that exploration in both tasks is difficult in the end-effector space.
To aid EE-Space exploration, we evaluated the approach from~\cite{pathak19disagreement}, which gives the agent additional exploration rewards.
While doing so leads the agent to cover a larger region in the state space, the explored states do not always correspond with meaningful behaviors for task completion, so we did not observe any gains using this method.

Unlike with the Block 2D tasks, 3-Priority is able to learn both the Franka tasks.
This is because the Franka tasks have fewer possible controllers, which resulted in lower dimensional priority-score action spaces.
The reduced action-space dimensions of Franka tasks allowed us to evaluate 3-Combo, which is also able to learn both tasks, although it achieves worse performance on Hex-Screw. 
Similarly, 1-Ctrlr is able make progress on Door-Open but not Hex-Screw, which suggests that Hex-Screw requires more precise coordination of multiple controllers than Door-Open.
Table~\ref{tab:franka_all_env_results} (rows 2, 3, 5 and 6) shows the success rates for both tasks on test configurations with small and large deviations.
All methods that use hierarchical combination of multiple object-axis controllers generalize well to both small and large test deviations.
Methods that performed poorly during training, EE-Space for both tasks and 1-Ctrlr Hex-Screw, do not generalize well. 

To evaluate Franka tasks in the real-world, we performed $10$ trials of each method on the real robot, each trial with a different sampled initial state.
For the Hex-Screw task, we further tested on $3$ different screw and key sizes.
All methods that used the proposed composition of hierarchical controllers were able to robustly perform Door-Open in the real world, while only 3-Exp-Single was able to do so for Hex-Screw.
Hex-Screw is more challenging than Door-Open, because it requires more precise movements for alignment and insertion.
As a result, sim-to-real gap in the robot dynamics and controller responses leads to greater performance degradation for Hex-Screw than for Door-Open.

%% file: includes/6_conclusion.tex
\section{Conclusion and Future Work}
\label{sec:conclusion}
\vspace{-3mm}

%

In this work, we propose using RL to learn how to compose hierarchical object-centric controllers for manipulation tasks.
Our approach has several advantages.
First, the object-centric controllers can be reused across multiple tasks.
Second, controller compositions are invariant to certain object properties.
Finally, the use of a structured action space introduces meaningful inductive biases for manipulation.
Our experiments show that the proposed approach leads to more guided exploration and consequently improved sample efficiency, and it enables zero-shot generalization to test environments and simulation-to-reality transfer without fine-tuning.
In future work, we will tackle the main limitations of the current approach -- the set of controllers is fixed and manually-defined.

%% file: includes/7_ack.tex
\acknowledgments{
\small{
This work was supported by NSF Award No. CMMI-1925130,
NSF Graduate Research Fellowship Program Grant No. DGE 1745016,
Office of Naval Research Grant No. N00014-18-1-2775, 
ARL grant W911NF-18-2-0218 as part of the A2I2 program,
and Nvidia NVAIL.
}
}
\normalsize

%% file: includes/8_appendix.tex
\begin{appendices}

\section{Controller Implementation Details}

\subsection{Specific Controllers for each Task}

\textbf{Block Fit}.
A set of controllers is associated with each wall in the environment.
For a wall, let $v$ be the unit vector pointing in the wall's normal direction, $x_{wm}$ be the coordinate of the middle of the wall.
The set of controllers associated for each wall include:
\begin{enumerate}
    \item Position attractor along normal direction. $x_d = x_{wm}$, $u = v$
    \item Position attractor along error direction. $x_d = x_{wm}$, $u = \frac{x_d - x_c}{\|x_d - x_c\|_2}$
    \item Force attractor along the normal direction. $f_d = 10$, $u = v$
    \item Rotation attractor aligning the block's x-axis to the normal. $r_d = v$, $u = [1, 0]^\top$
    \item Rotation attractor aligning the block's y-axis to the normal. $r_d = v$, $u = [0, 1]^\top$
\end{enumerate}

\textbf{Block Push}.
In addition to all the per-wall controllers of the Block Fit task, Block Push has the following per-wall controllers:
\begin{enumerate}
    \item Position controller along the side of a wall.
    Let $v'$ be a unit vector orthogonal to $v$.
    Since there are $2$ such possible directions, we pick the one that gives the direction pointing up along the vertical wall in the scene.
    
    Let $x_{wc}$ be the coordinate of a wall corner. Since walls form a corner-connect chain in this task, using one of the two corners per wall covers all corners in the scene except the last corner in the chain, which we ignore.
    
    With these, this controller has $x_d = x_{wc}$ and $u = v'$.
    
    \item Position curl controller around a wall corner.
    This controller rotates the end-effector in a fixed-radius circle around a point until it reaches the target position which also lies on the circle.
    The attractor target is $x_d = x_{wc} + \|x_c - x_{wc}\|_2 v'$, and the direction is $u = R(\frac{\pi}{2}) \frac{x_c - x_{wc}}{\|x_c - x_{wc}\|_2}$, where $R(\theta)$ gives a 2D rotation matrix with the angle $\theta$.
\end{enumerate}

Block Push has one more position controller that attracts the robot block toward the target block.
Let $x_g$ be the current location of the center of the target block.
This position controller has $x_d = x_g$ and $u = \frac{x_d - x_c}{\|x_d - x_c\|_2}$.

\textbf{Franka Hex-Screw}.
Let $x_s$ be the location of screw, and $x_g = x_s + [0, 0.02, 0]^\top$ be a point $2$cm above the screw (the $y$-axis is vertical in our coordinate frame).
Position attractor controllers use $x_g$ as the target, instead of $x_s$, because attracting the hex-key tip toward the inside of the screw directly can result in collisions with the side of the screw and prevent the key from properly inserted.
\begin{enumerate}
    \item Position attractor along vertical direction. $x_d = x_g$, $u = [0, 1, 0]^\top$
    \item Position attractor along error direction. $x_d = x_g$, $u = \frac{x_d - x_c}{\|x_d - x_c\|_2}$
    \item Position controller that prevents motion in the vertical direction $x_d = x_c$, $u = [0, 1, 0]^\top$.
    This controller does not attract the end-effector toward a goal.
    Instead, its utility is solely in its nullspace projection, which ensures lower-priority controllers cannot move the end-effector outside of a horizontal plane.
    This controller is useful for preventing prematurely inserting the hex-key.
    \item Force controller that pushes downward toward the hex screw. $f_d = 20$, $u = [0, -1, 0]^\top$.
    \item Rotation controller that maintains the verticality of the end-effector. $r_d = [0, 1, 0]^\top$, $u = [0, 0, 1]^\top$. The positive $z$-axis of the end-effector frame corresponds to the direction that the hex-key points towards.
    \item Rotation controller that rotates the hex-key counter-clockwise. $r_d = R_y(100^\circ) [1, 0, 0]^\top$, $u = [1, 0, 0]^\top$, where $R_y(\theta)$ gives a rotation matrix that rotates around the $y$-axis with the angle $\theta$.
    \item Rotation controller that rotates the hex-key clockwise. $r_d = R_y(-100^\circ) [1, 0, 0]^\top$, $u = [1, 0, 0]^\top$
\end{enumerate}

\begin{figure}[!t]
    \centering
    \includegraphics[width=0.3\linewidth]{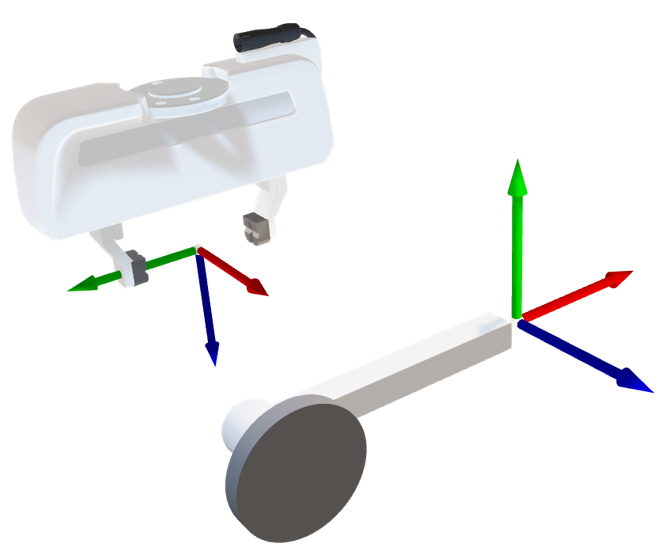}
    \caption{
    \footnotesize{
    Axes Visualization for Franka End-Effector and Door Handle for the Door-Open Task. RGB corresponds to XYZ.
    }}
    \label{fig:franka_door_axes}
\end{figure}

\textbf{Franka Door-Open}.
See Figure~\ref{fig:franka_door_axes} for a visualization of both the Franka end-effector and door handle axes.
Let $r_{\{x, y, z\}}$ correspond to the $3$ axes of the door handle.
Let $x_g$ be a grasp point on the door handle, $x_h$ be the center of the handle axle (dark gray cylinder in Figure~\ref{fig:franka_door_axes}).
The set of controllers include:
\begin{enumerate}
    \item Position attractor to door handle along error direction. $x_d = x_g$, $u = \frac{x_d - x_c}{\|x_d - x_c\|_2}$
    \item Position curl attractor for rotating around the handle in the plane of the door panel (the nullspace of $r_z$). Let $x_e = \mathcal{N}(r_z)(x_c - x_h)$. Then $x_d = x_h - \|x_e\|_2 r_y$, 
    $u = \frac{x_e}{\|x_e\|_2} \times r_z$
    \item Force controller to pull the handle. $f_d = 50$, $u = -r_z$
    \item Rotation controller to align the x-axes of the gripper and the handle. $r_d = r_x$, $u = [1, 0, 0]^\top$
    \item Rotation controller to align the y-axes of the gripper and the handle. $r_d = r_y$, $u = [0, 1, 0]^\top$
    \item Rotation controller to align the z-axes of the gripper and the handle. $r_d = r_z$, $u = [0, 0, 1]^\top$
\end{enumerate}

\subsection{Integral Term for Force Controllers}
Using an integral term for force controllers can help reduce the force error and improve stability.
Let $\bar{\delta_f^i}$ be the accumulated force errors for the force controller used at the $i$th priority.
Then, the corresponding delta position target is computed as:
\begin{equation}
    \Delta_x^i = \mathcal{N}_i (K_f \delta_f(f_d^i, u^i, f_c) + K_I \bar{\delta_f^i})
\end{equation}
where $\mathcal{N}_i = \mathcal{N}([u_0,\hdots,u_{i-1}])$.

\subsection{Delta Target Magnitude Clipping}
To ensure safety and limit the maximum speed at which our controllers can drive the robot, we clip the magnitude of delta position and rotation targets.

Let $D_x$ be the maximum delta translation magnitude corresponding to a position controller, and $D_f$ for a force controller.
The clipping for force and position controllers are computed as follows:
\begin{align}
    \Delta_x^i \leftarrow \frac{\min(\|\Delta_x^i\|_2, D_*)}{\|\Delta_x^i\|_2} \Delta_x^i
\end{align}
Note that $D_*$ can be $D_x$ or $D_f$, depending on if the $i$th controller is a position or force controller.

Similarly, let $D_R$ be the maximum delta rotation angle for rotation controllers:
\begin{align}
    \Delta_R^i \leftarrow \frac{\min(\|\Delta_R^i\|_2, D_R)}{\|\Delta_R^i\|_2} \Delta_R^i
\end{align}

\subsection{Controller Hyperparameters}

Table~\ref{tab:hparams_controllers} lists the different hyperparameters used for the object axes-controllers for each task. We list the gains used for each controller as well as the clipping used while executing each controller. Table~\ref{tab:hparams_impedance_control} lists the task-space impedance parameters used for simulation and real-world experiments. 
We use \cite{zhang2020modular} to implement each controller for real-world experiments.

\begin{table}[!h]
\centering
\begin{tabular}{l|llll}
\toprule
            & Block Fit & Block Push & Franka Hex-Screw & Franka Door-Open \\ \midrule
$D_x$ (m)   & $1$       & $0.5$      & $0.03$           & $0.03$           \\
$D_f$ (N)   & $0.5$     & $0.1$      & $10^{-4}$        & $10^{-4}$        \\
$D_R$ (deg) & $90$      & $120$      & $10$             & $10$             \\ \midrule
$K_x$       & $1$       & $1$        & $1$              & $1$              \\
$K_f$       & $1$       & $1$        & $1$              & $1$              \\
$K_I$       & $0$       & $0$        & $10^{-4}$        & $10^{-4}$        \\ \bottomrule
\end{tabular}
\vspace{5pt}
\caption{\footnotesize{Controller Gains and Magnitude Clips Across Tasks.}}
\label{tab:hparams_controllers}
\end{table}

\begin{table}[!h]
\centering
\begin{tabular}{l|ll}
\toprule
      & Simulation     & Real World    \\ \midrule
$K_S$ & $1000$         & $600$         \\
$K_D$ & $2\sqrt{1000}$ & $2\sqrt{600}$ \\
$T$   & $10$           & $30$           \\ \bottomrule
\end{tabular}
\vspace{5pt}
\caption{\footnotesize{Task-Space Impedance Control Parameters. $K_S$ is stiffness, $K_D$ is damping, and $T$ is how many timesteps a controller combination runs before the RL policy is queried again. The simulation and real-world values are not the same due to differences in control frequencies and Franka dynamics between real-world and simulation. We tune the real-world values to ensure that the resultant controller behaviors are similar to those in simulation. This tuning was done prior to task evaluations.}}
\label{tab:hparams_impedance_control}
\end{table}

\section{Task Details}

\subsection{Block Fit}

\textbf{Observations}.
\begin{enumerate}
    \item 2D pose of block robot
    \item 2D contact force direction and magnitude experienced by the block robot in the world frame.
    \item 2D coordinates of centers and wall corners
\end{enumerate}

\textbf{Reward Function}:
Let $\phi_d$ be the previous distance between the block translation and the goal translation, $\phi_\theta$ be previous the absolute angle difference between the block rotation and the goal rotation, and let $\phi'_d$, $\phi'_\theta$ be there current counterparts.
The reward function rewards making progress towards the goal with a small alive penalty and a large task completion bonus:
\begin{equation}
    R = 10 (\phi'_d - \phi_d) + 5(\phi'_\theta - \phi_\theta) - 0.1 + 1000 \times\mathbbm{1}(\phi'_d < 0.16 \wedge \phi'_\theta < 5^\circ)
\end{equation}
The goal translation threshold $0.16$ is about half the width of the block.


\subsection{Block Push}

\textbf{Observations}.
\begin{enumerate}
    \item 2D pose of block robot
    \item 2D contact force direction and magnitude experienced by the block robot in the world frame.
    \item 2D coordinates of centers and wall corners
    \item 2D pose of the target block
\end{enumerate}

\textbf{Reward Function}:
The reward function is similar to that of Block Fit, except the progress rewards are computed w.r.t. the target block, not the block robot, and there is an additional reward term for approaching the target block:
\begin{equation}
    R = 10 (\phi'_d - \phi_d) + 10 (\phi'_b - \phi_b) - 0.1 + 200 \times\mathbbm{1}(\phi'_d < 0.05)
\end{equation}
where $\phi_b$ is the previous distance between the robot block and the target block, and $\phi'_b$ is the current counterpart.
The goal translation threshold of $0.05$ is about half the width of the target block.

\subsection{Franka Hex Screw}

\begin{figure}[!t]
    \centering
    \includegraphics[width=0.35\linewidth]{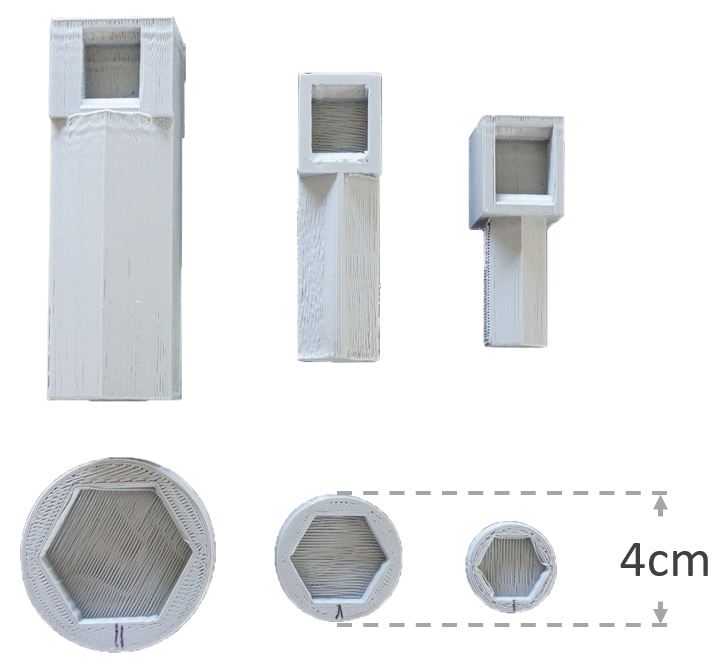}
    \caption{\footnotesize{
        Different hex screw and key sizes used for testing in the real world.
        The middle size represents $1.0\times$ scale factor, while the left is $1.5\times$, and the right $0.7\times$.
    }}
    \label{fig:hex_sizes}
\end{figure}

See Figure~\ref{fig:hex_sizes} for the different screw and key sizes used in real-world experiments.

\textbf{Observations}.
\begin{enumerate}
    \item 7-dimension robot arm joint angles
    \item Gripper width
    \item 6D pose of the tip of the hex-screw. Rotations are represented via quaternions
    \item End-effector contact forces
    \item Position of the hex screw relative to the robot base
\end{enumerate}

\textbf{Reward Function}:
Let $\phi_d$ be the previous distance from the hex-key tip to the hex screw, $\phi_\theta$ be the previous absolute angle difference between the screw angle and its target angle, and let $\phi'_d$, $\phi'_\theta$ be their respective current counterparts.
Let $\rho$ be the absolute angle difference between the negative $y$-axis (pointing downwards) and the $z$-axis of the end-effector.
The reward function rewards approaching the hex-screw, making progress in turning the screw, maintaining a vertical hex key orientation, plus a small alive penalty and a large task bonus:
\begin{equation}
    R = 400(\phi'_d - \phi_d) + 10(\phi'_\theta - \phi_\theta) - \rho - 1 + 1000\times\mathbbm{1}(\phi'_\theta < 5^\circ)
\end{equation}
The target screw rotation angle (at which point $\phi_\theta=0$) is $70^\circ$.

\subsection{Franka Door Opening}

\textbf{Observations}.
\begin{enumerate}
    \item 7-dimension robot arm joint angles
    \item Gripper width
    \item 6D pose of the tip of the hex-screw. Rotations are represented via quaternions
    \item End-effector contact forces
    \item Door panel angle (How much the door has opened, not the angle of the door handle)
    \item Position of the door handle relative to the robot base
\end{enumerate}

\textbf{Reward Function}:
Let $\phi_d$ be the previous distance from the end-effector to the door handle grasp point, $\phi_\theta$ be the previous absolute angle difference between the door handle angle and the target handle turning angle, $\phi_\rho$ be the previous absolute angle difference between the door panel angle and the target door opening angle, and let $\phi'_d$, $\phi'_\theta$, $\phi'_\rho$ be their respective current counterparts.
Let $c$ denote the current end-effector contact forces.
The reward function rewards approaching the door handle, turning the handle, turning the door, plus small alive penalties and excessive force penalties, plus a large task bonus:
\begin{equation}
    R = 10(\phi'_d - \phi_d) + 10(\phi'_\theta - \phi_\theta) + 100(\phi'_\rho - \phi_\rho)  - 0.01 - 0.001 \|c\|_2 + 100\times\mathbbm{1}(\phi'_\rho < 5^\circ)
\end{equation}
The target door handle turning angle (at which point $\phi_\theta=0$) is $90^\circ$, and the target door panel opening angle (at which point $\phi_\rho=0$) is $35^\circ$.


\section{RL Training Details}

\subsection{PPO Hyperparameters}
\begin{table}[!h]
\centering
\begin{tabular}{l|llll}
\toprule
                           & Block Fit           & Block Push          & Franka Hex-Screw    & Franka Door-Open    \\
\midrule
num steps                  & $500$               & $240$               & $450$               & $480$               \\
discount factor            & $0.995$             & $0.995$             & $0.995$             & $0.995$             \\
entropy coefficient        & $0.01$              & $0.01$              & $[0.01, 0.1]$       & $[0.01, 0.1]$              \\
learning rate              & $2.5\times 10^{-4}$ & $2.5\times 10^{-4}$ & $2.5\times 10^{-4}$ & $2.5\times 10^{-4}$ \\
value loss coefficient     & $0.5$               & $0.1$              & $0.5$               & $0.5$               \\
max gradient norm          & $0.5$               & $0.5$               & $0.5$               & $0.5$               \\
lambda                     & $0.95$              & $0.95$              & $0.95$              & $0.95$              \\
num minibatches            & $50$                & $30$                & $30$                & $50$                \\
num opt epochs    & $4$                 & $4$                 & $4$                 & $4$                 \\
clip range          & $0.2$               & $0.2$               & $0.2$               & $0.2$               \\
\bottomrule
\end{tabular}
\vspace{5pt}
\caption{\footnotesize{PPO Hyperparameters Across All Tasks.}}
\label{tab:hparams_ppo}
\end{table}

Table~\ref{tab:hparams_ppo} lists the hyperparameters used for each of the experiments. 
In addition to the above parameters, we also decay the 
clip range using a linear schedule with a decay rate of 0.99 after every epoch. We set the minimum clip range value to be $0.04$.
Also, for the Franka Hex-Screw and Franka Door-Open task we evaluated a range of entropy coefficient values $[0.01, 0.1]$ for the end-effector action space.

\subsection{Network Architecture}
For all tasks and methods, we use the same network architectures for both the policy and value function networks.
The network consists of 2 hidden layers with $256$ hidden units each.

\subsection{Controller Features in the Observation Space}

\begin{figure}[!ht]
    \centering
    \includegraphics[width=0.9\linewidth]{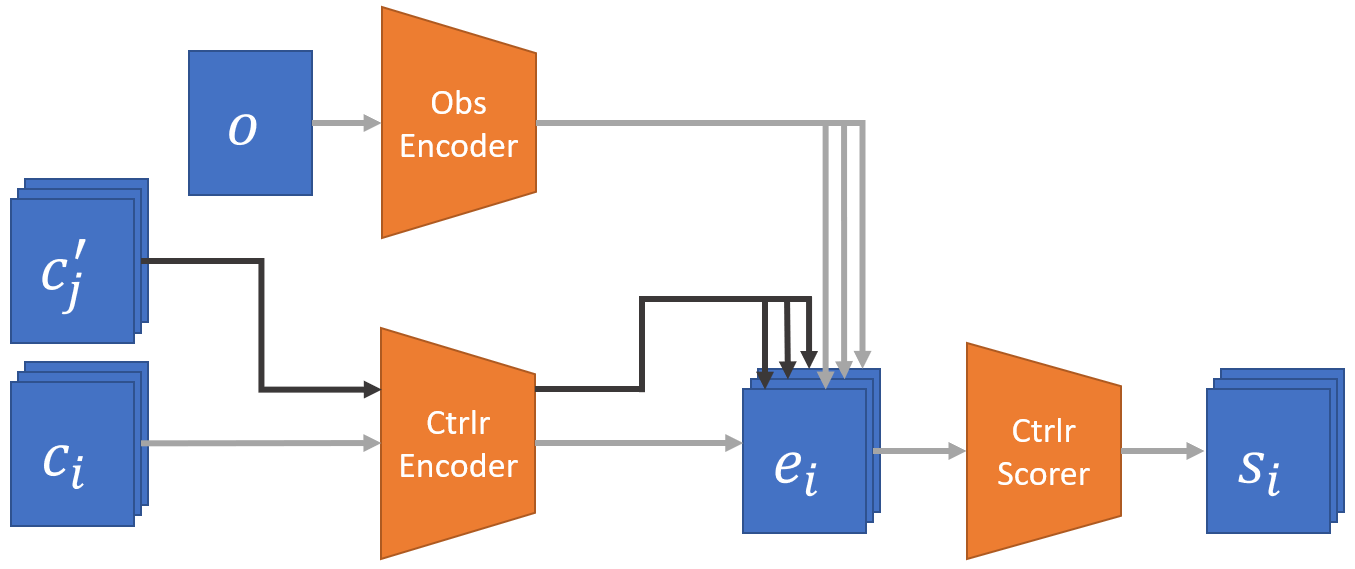}
    \caption{Policy Architecture for 3-Exp-Features.}
    \label{fig:ctrlr_features}
\end{figure}

We experimented with giving features of each individual controllers to the RL policy in the Expanded-MDP approach.
These features may allow the policy to better reason about the effects of individual controllers, and it also allows the policy to operate on a variable number of controllers.
Controller features include a 1-hot encoding of the controller type (position, force or rotation), the controller target, axis, and the current error.
We refer to this method as \textbf{3-Exp-Features}.

See Figure~\ref{fig:ctrlr_features} for an illustration of a policy architecture using controller features.
Each controller feature vector $c_i$ of the $i$th controller (there are $N$ in total) is processed by a shared controller feature encoder.
These controller embeddings are then each concatenated with embeddings of the original observations, which include environment observations $o$ and encodings of previously selected controllers $c'_j$ (there are $N_c - 1$ in total).
For $c'_j$, instead of single-1-hot or multi-1-hot embeddings, we also use the controller features, which are first processed by the shared controller encoder.
Finally, the concatenated embeddings $e_i$ are separately processed and scored by the policy.
The normalized scores $s_i$'s form the discrete distribution from which we sample the next controller selection.

Both the observation and controller feature encoders contain two hidden layers of $64$ hidden units each.
The controller scorer has one hidden layer of $32$ hidden units.
For the value function, we remove the last linear layer (size $32\times 1$) of the controller scorer in the policy, add the $32$-dimensional outputs from the hidden layer across all $N$ intermediate outputs, and pass the sum through one last linear layer (also of size $32\times 1$) to obtain one scalar value.

While 3-Exp-Features policies are invariant to the number of controllers in the scene, we did not explicitly test for this capability.
However, we still ran this method alongside the other approaches, and it achieves comparable performance to the other 3-Exp variants (see detailed results below).

\section{Detailed Experiment Results}

We discuss results for each task in more detail in the following sections. 
Video results for all the different tasks and methods can be seen at \url{https://sites.google.com/view/compositional-object-control/}.

\subsection{Block Fit}

\begin{figure}
\centering
\begin{subfigure}[b]{0.95\textwidth}
   \includegraphics[width=1\linewidth]{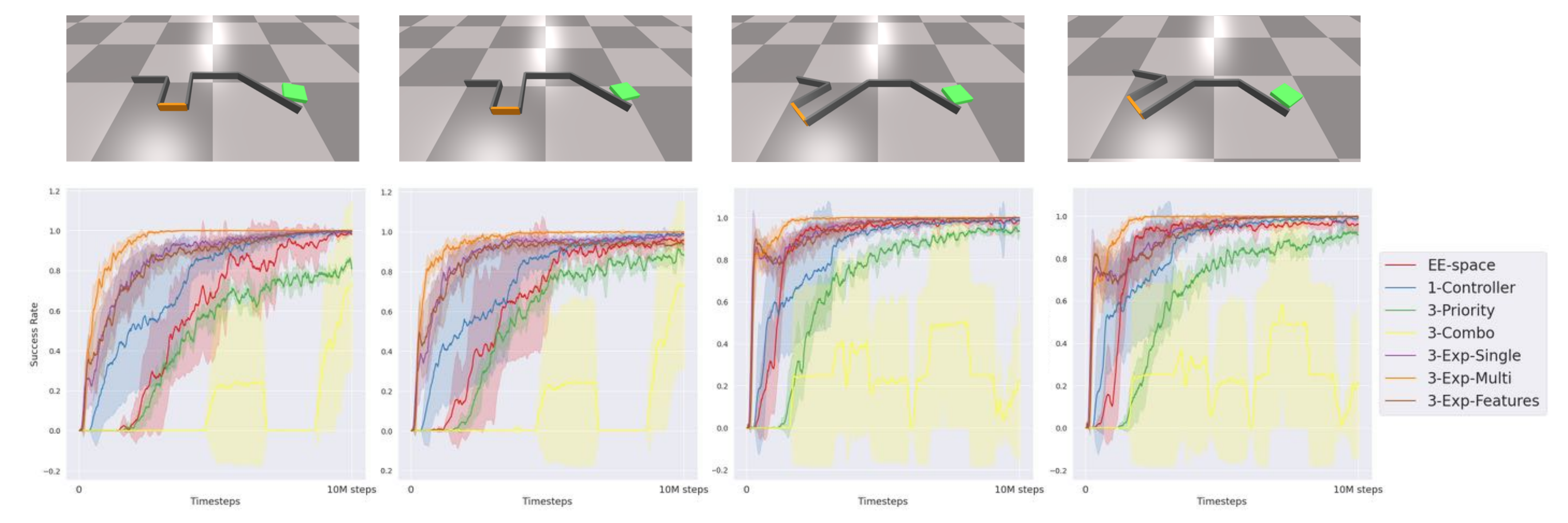}
   \caption{}
   \label{fig:block2d_train_1} 
\end{subfigure}

\begin{subfigure}[b]{0.95\textwidth}
   \includegraphics[width=1\linewidth]{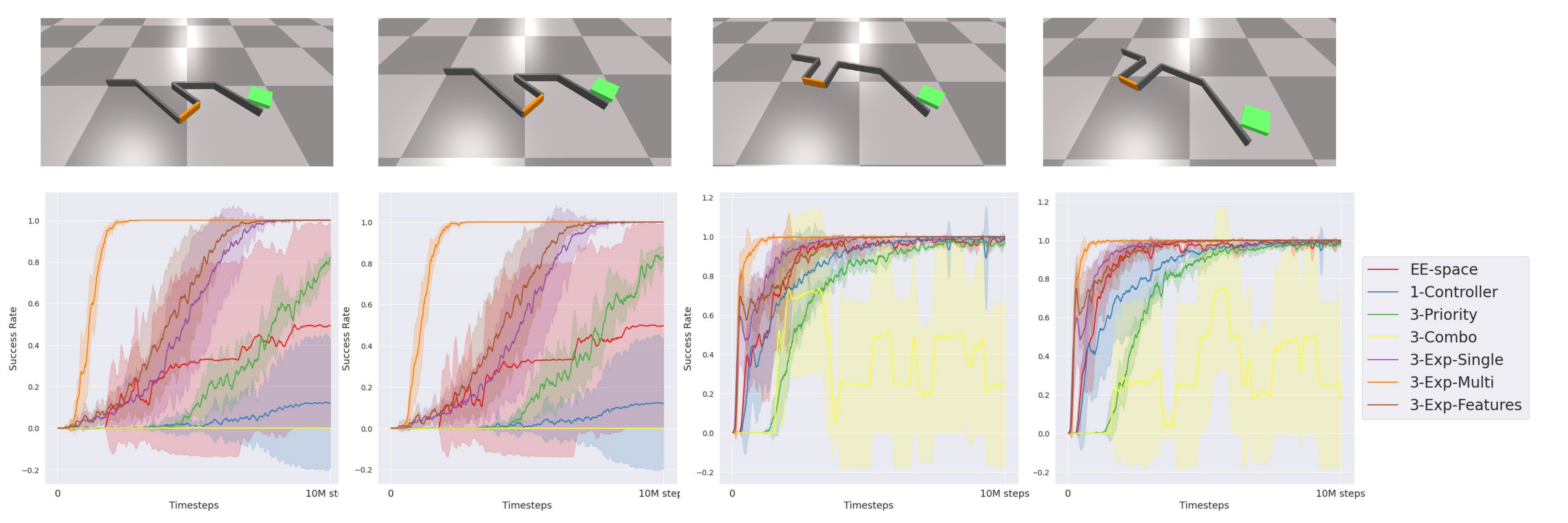}
   \caption{}
   \label{fig:block2d_train_2}
\end{subfigure}
\caption{Different environment configurations used to train the Block Fit task. The plot below each environment configuration shows how the trained policy performed on each particular configuration.}
\label{fig:block2d_fit_train_results}
\end{figure}

\begin{table}[h]
\begin{minipage}{0.98\textwidth}
\centering
    \begin{figure}[H]
    \begin{subfigure}[b]{0.95\textwidth}
       \includegraphics[width=1\linewidth]{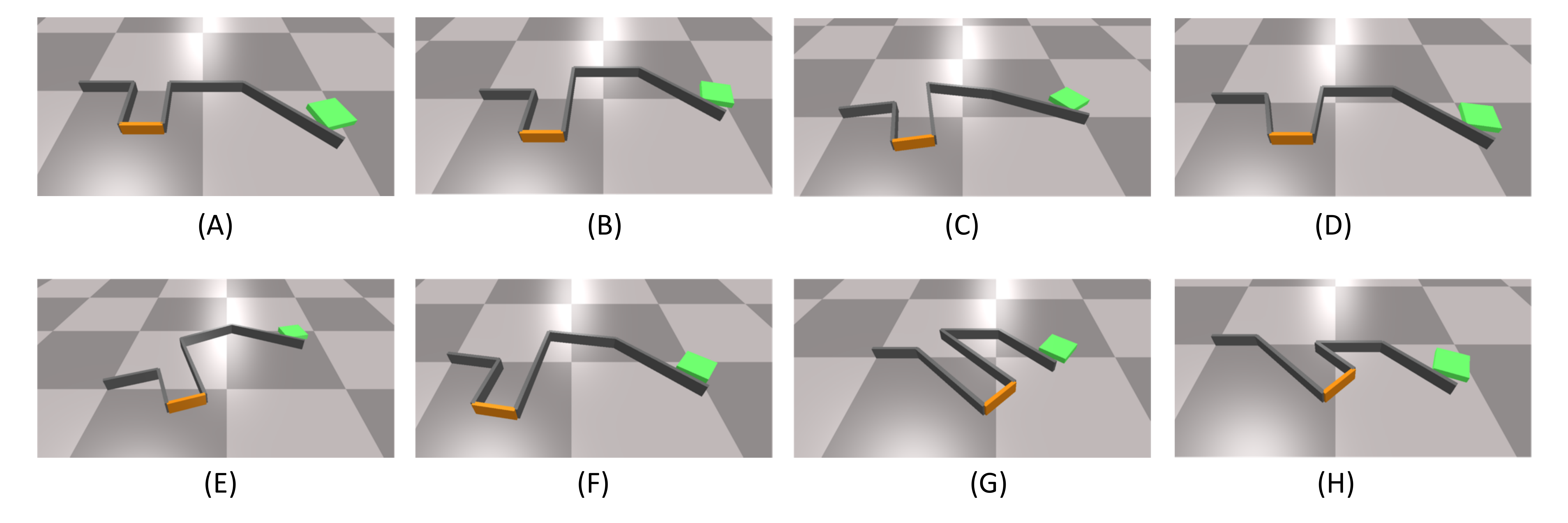}
       \caption{}
       \label{fig:block2d_test_all} 
    \end{subfigure}
    \caption{Test configurations for the Block Fit task. Table~\ref{tab:block2d_fit_test_results} shows results on each environment configuration.} 
    \end{figure}
\end{minipage} \\
\begin{minipage}{1.0\textwidth}
    \begin{table}[H]
    \resizebox{1.0\textwidth}{!}{%
    \begin{tabular}{@{}llllllll@{}}
    \toprule
    Test-Cfg & EE-Space & 1-Ctrl & 3-Pri. & 3-Combo & 3-Exp-Feat. & 3-Exp-Single & 3-Exp-Multi \\ \midrule
    A & 0.86 \small{(0.07)} & 0.98 \small{(0.01)} & \textbf{1.0 \small{(0.0)}} & 0.19 (0.12) & \textbf{1.0 \small{(0.0)}}  & \textbf{1.0 \small{(0.0)}} & \textbf{1.0 \small{(0.0)}} \\
    B & 1.0 \small{(0.0)} & 1.0 \small{(0.0)} & \textbf{1.0 \small{(0.0)}}  & 0.20 (0.14) & \textbf{1.0 \small{(0.0)}}  & \textbf{1.0 \small{(0.0)}} & \textbf{1.0 \small{(0.0)}} \\
    C & 0.85 \small{(0.27)} & 0.96 \small{(0.03)} & 0.97 \small{(0.03)} & 0.18 (0.08) &  0.98 (0.01) & \textbf{0.99 \small{(0.01)}} & 0.96 \small{(0.01)} \\
    D & 0.89 \small{(0.14)} & 0.71 \small{(0.31)} & \textbf{1.0 \small{(0.03)}} &  0.165 (0.06)  & \textbf{1.0 \small{(0.0)}} & \textbf{1.0 \small{(0.0)}} & \textbf{1.0 \small{(0.01)}} \\
    E & 0.16 \small{(0.23)} & 0.64 \small{(0.16)} & 0.76 \small{(0.16)} &  0.0375 (0.05) & 0.97 (0.02) & \textbf{0.99 \small{(0.01)}} & 0.99 \small{(0.02)} \\
    F & 0.75 \small{(0.39)} & 0.87 \small{(0.16)} & \textbf{1.0 \small{(0.0)}} & 0.20 (0.14) & 0.98 (0.01)  &  \textbf{1.0 \small{(0.0)}} & \textbf{1.0 \small{(0.0)}} \\
    G & 0.17 \small{(0.30)} & 0.08 \small{(0.23)} & 0.75 \small{(0.45)} & 0.02 (0.03)  & 0.89 (0.13) &  0.82 \small{(0.23)} & \textbf{0.90 \small{(0.15)}} \\
    H & 0.50 \small{(0.53)} & 0.0 \small{(0.0)} & \textbf{1.0 \small{(0.0)}} & 0.27 (0.18) & \textbf{1.0 \small{(0.0)}} &  \textbf{1.0 \small{(0.0)}} & \textbf{1.0 \small{(0.0)}} \\
    \bottomrule \\
    \end{tabular}%
    }
    \caption{Block Fit mean success on test environment configurations. Parentheses denote standard deviation across 8 seeds.} 
    \label{tab:block2d_fit_test_results}
    \end{table}
\end{minipage}
\end{table}



Figure~\ref{fig:block2d_fit_train_results} plots the train results (success-rate) for all different train configurations. 
As can be seen in the above figure, for all configurations besides a couple (Figure~\ref{fig:block2d_train_2} left two plots), all the methods perform quite well during training. 
The poor performance in two of the training environments is due to their more challenging configurations. 
In both of these configurations the slot to the target wall is oriented in a different direction, so a robust policy needs to reason about this change. 
We observe that our proposed Expanded-MDP methods are able to perform well for this configuration. 
However, the end-effector action space shows a large variance \emph{i.e.} many of the seeds fail to learn a robust policy to solve these configurations. 
Additionally, we also observe that 1-Ctrlr is unable to solve this task robustly. 
This shows the advantage of using multiple-controllers in parallel.

Figure~\ref{fig:block2d_test_all} shows the different test configurations we evaluated. Each of the test configurations is a delta change in the wall lengths or angles from the train configurations.
Table~\ref{tab:block2d_fit_test_results} shows the results on each of these test configurations. 
As seen above, our proposed Expanded-MDP formulations are able to outperform all other methods for all the configurations. 
3-Priority performs well on most test configurations besides the slightly harder ones (E, G).
This indicates that Expanded-MDP methods are able to learn more robust policies as compared to using a continuous priority score. 
Additionally, EE-Space perform poorly, especially for more different test configurations (E, F, G, H).
Qualitatively, we observe that EE-Space policies often completely fail to generalize to the test configurations. 
1-Ctrlr performs poorly on the D, E, and G, H configurations. 
For the initial two configurations, we observe that the learned policy can often get stuck around wall corners, which prevents it from completing the episode within the given time. 
Alternately, for the latter two environments, the learned policies across all seeds perform quite poorly, so they are not able to generalize to either of the test configurations.

\subsection{Block Push}

\begin{figure}
\centering
\begin{subfigure}[b]{0.95\textwidth}
   \includegraphics[width=1\linewidth]{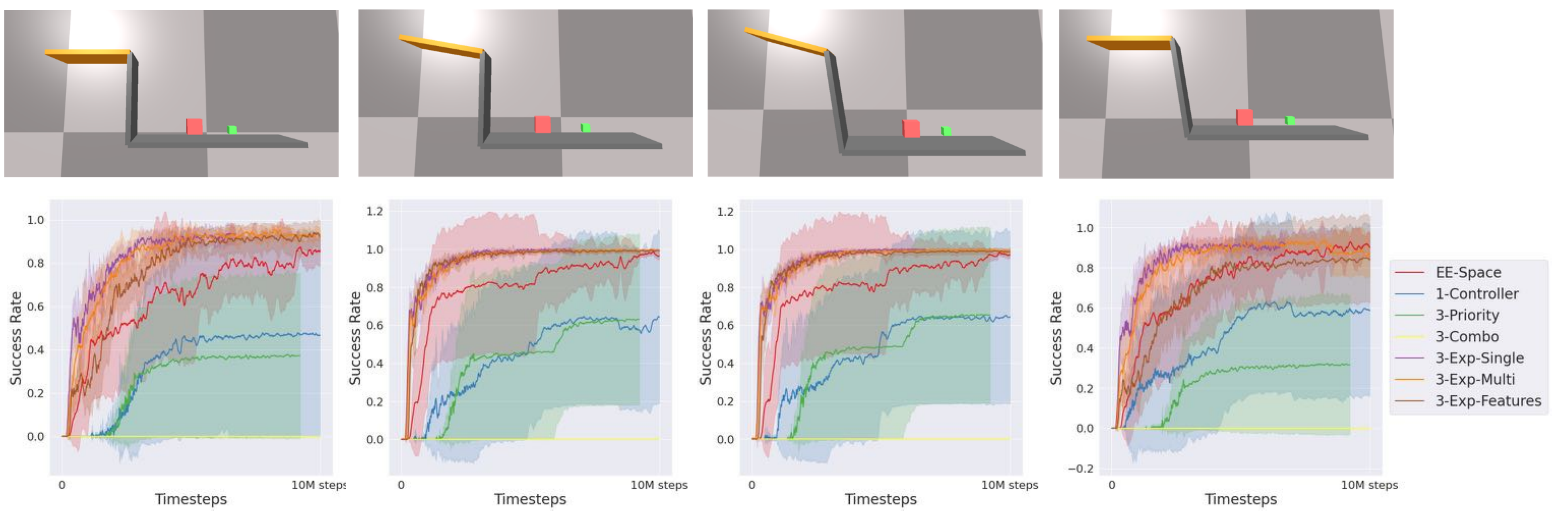}
   \caption*{}
   \label{fig:push_block_train_1} 
\end{subfigure}

\begin{subfigure}[b]{0.95\textwidth}
   \includegraphics[width=1\linewidth]{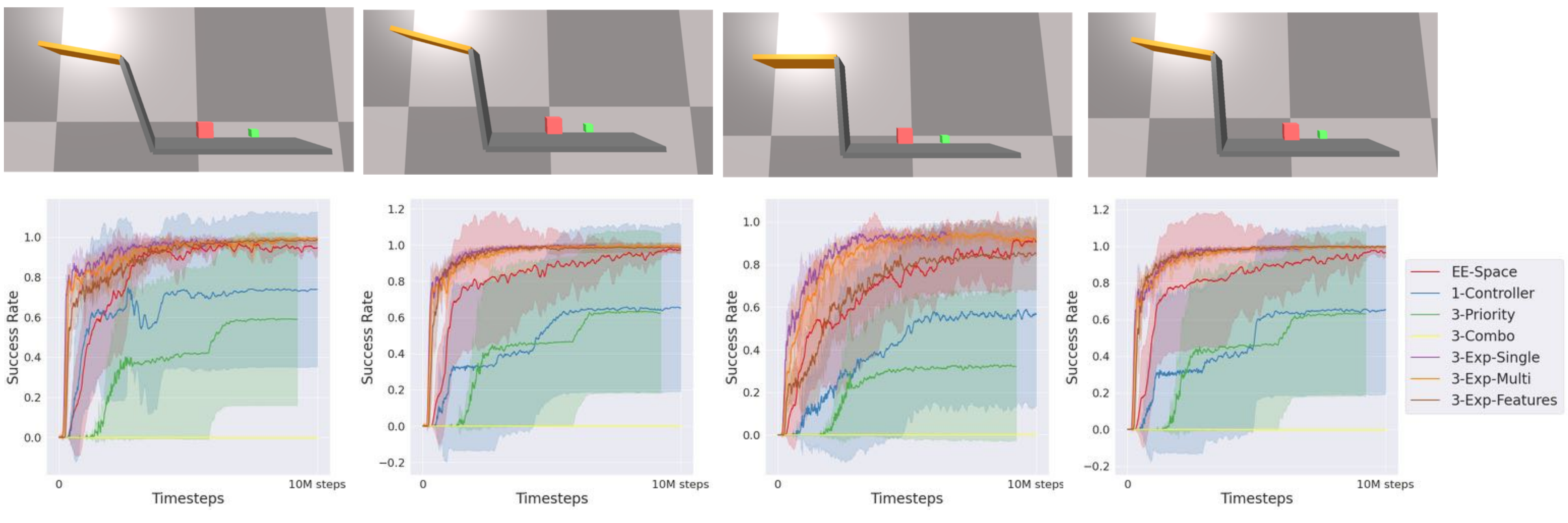}
   \caption*{}
   \label{fig:push_block_train_2} 
\end{subfigure}

\begin{subfigure}[b]{0.95\textwidth}
   \includegraphics[width=1\linewidth]{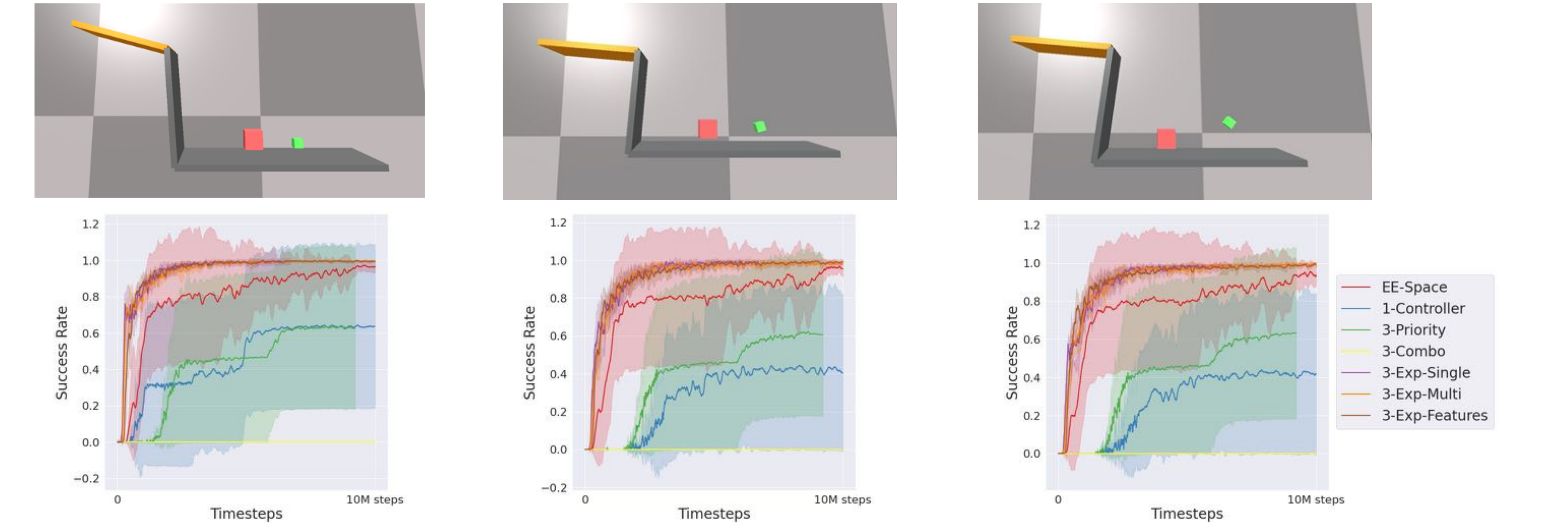}
   \caption*{}
   \label{fig:push_block_train_3} 
\end{subfigure}

\caption{Different environment configurations used to train the Block Push task. The plot below each environment configuration shows how the trained policy performed on each particular configuration.}
\label{fig:block2d_train_results_push}
\end{figure}

Figure~\ref{fig:block2d_train_results_push} shows the success rate for all the different environment configurations used in the Block Push task. 
As seen in the above figure, both 1-Ctrlr and 3-Priority show large variance in training performance.
This is because both methods fail to learn the task for some seeds.
All the Expanded-MDP methods are able to successfully complete the task without large variance. 
One reason for this is that a robust task strategy requires the use of multiple object-axis controllers to move the object along the vertical wall as well as to move it around the corner of the top wall. 
Although it is feasible to accomplish the task by quickly switching between controllers, it is hard to find a robust policy relying under such a switching mechanism, especially when controllers are being run for fixed number of steps.
Additionally, while EE-Space also solves all the different environment configurations, its sample complexity is worse than the proposed Expanded-MDP methods. 

Table~\ref{tab:block2d_push_gen_result} evaluates the learned policy on $8$ different test configurations. 
Figure~\ref{fig:push_block2d_test_all} plots each of these test configurations. 
These test configurations involve small perturbations in either the wall length or the wall angles (or both) from the train configurations.
Specifically, we limit small perturbations to be a max change in vertical wall angle of $3-5^\circ$ (A, C, D, F), while larger perturbations are sampled from $\sim 6-10^\circ$ (B, E, G, H).
We observe that EE-Space is usually robust to small perturbations of the environment, while slightly larger perturbations can significantly affect its performance. 
However, even with small perturbations our expand-MDP based methods are able to outperform the end-effector space.
This shows the advantage of using a structured action space for learning, as Expanded-MDP methods perform well across both sets of configurations. 
Notably, the proposed approach only performs poorly on B and E configurations. 
For both configurations, as the policy pushes the red block up the middle wall, the agent block (green block) can sometimes end up under the red block, which leads to the green block falling, ending the episode.
Additionally, both 1-Ctrlr and 3-Priority perform poorly on the test configurations. 
This is due to the poor performance of some of the learned policies (across a few seeds) on the train configurations. 
However, good train performance does not necessarily lead to good test performance.
Specifically, 1-Ctrlr can move the green block upwards (by using the position controller for the top-wall), but it is often not able to robustly push it around the corner. 
This shows the advantage of being able to choose multiple object-axis controllers at each step.


\begin{table}[h]
\begin{minipage}{0.98\textwidth}
\centering
    \begin{figure}[H]
    \begin{subfigure}[b]{0.95\textwidth}
       \includegraphics[width=1\linewidth]{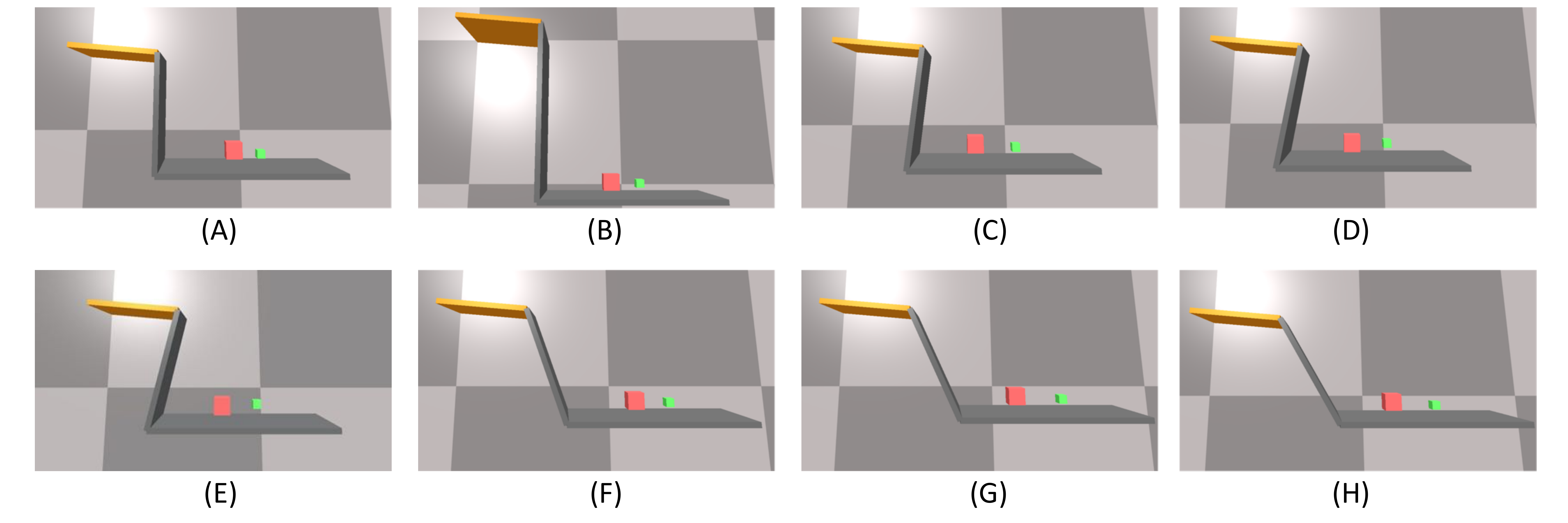}
       \caption{}
       \label{fig:push_block2d_test_all} 
    \end{subfigure}
    \caption{Test environment configurations for the Block Push task. Table~\ref{tab:block2d_push_gen_result} shows results on each environment config.} 
    \end{figure}
\end{minipage} \\
\begin{minipage}{0.98\textwidth}
    \begin{table}[H]
    \resizebox{\textwidth}{!}{%
    \begin{tabular}{@{}llllllll@{}}
    \toprule
    Config & EE-Space & 1-Ctrl & 3-Priority & 3-Combo  & 3-Exp-Feat. & 3-Exp-Single & 3-Exp-Multi \\ \midrule
    A & 0.94 \small{(0.06)} & 0.62\small{(0.47)} & 0.43 \small{(0.47)} & 0.0 (0.0)  & 0.97 \small{(0.02)}  & 0.98 \small{(0.01)} & \textbf{0.99 \small{(0.00)}} \\
    B & 0.27 \small{(0.28)} & 0.27\small{(0.30)} & 0.38 \small{(0.43)} & 0.0 (0.0)  & 0.50 \small{(0.27)}  & \textbf{0.72 \small{(0.18)}} & 0.65 \small{(0.30)} \\
    C & 0.86 \small{(0.23)} & 0.30 \small{(0.40)} & 0.43\small{(0.37)} & 0.0 (0.0)  & 0.91 \small{(0.10)}  & \textbf{0.97 \small{(0.01)}} & \textbf{0.97 \small{(0.04)}} \\
    D & 0.70 \small{(0.28)} & 0.07 \small{(0.13)} & 0.42 \small{(0.46)} & 0.0 (0.0)  & 0.89 \small{(0.06)} & \textbf{0.93 \small{(0.07)}} & 0.88 \small{(0.12)} \\
    E & 0.48 \small{(0.31)} & 0.01 \small{(0.01)} & 0.36 \small{(0.39)} & 0.0 (0.0)  & \textbf{0.79 \small{(0.17)}} & 0.69 \small{(0.23)} & 0.67 \small{(0.17)} \\
    F & \textbf{0.96 \small{(0.03)}} & 0.73 \small{(0.41)} & 0.38 \small{(0.42)} & 0.0 (0.0)  & 0.88 \small{(0.11)} & 0.95 \small{(0.06)} & \textbf{0.96 \small{(0.03)}} \\
    G & 0.89 \small{(0.10)} & 0.67 \small{(0.49)} & 0.35 \small{(0.39)} & 0.0 (0.0)  & \textbf{0.97 \small{(0.03)}} & 0.92 \small{(0.06)} & 0.89 \small{(0.07)} \\
    H & 0.61 \small{(0.26)} & 0.27 \small{(0.41)} & 0.34 \small{(0.38)} & 0.0 (0.0)  & 0.79 \small{(0.11)} & 0.78 \small{(0.10)} & \textbf{0.88} \small{(0.07)} \\
    \bottomrule \\
    \end{tabular}%
    }
    \caption{Block Push mean success on test environment configurations. Parentheses denote standard deviation across 8 seeds.} 
    \label{tab:block2d_push_gen_result}
    \end{table}
\end{minipage}
\end{table}

\subsection{Franka Hex-Screw}

\begin{figure}[ht]
\begin{subfigure}{.55\textwidth}
  \centering
  \includegraphics[width=1.0\linewidth]{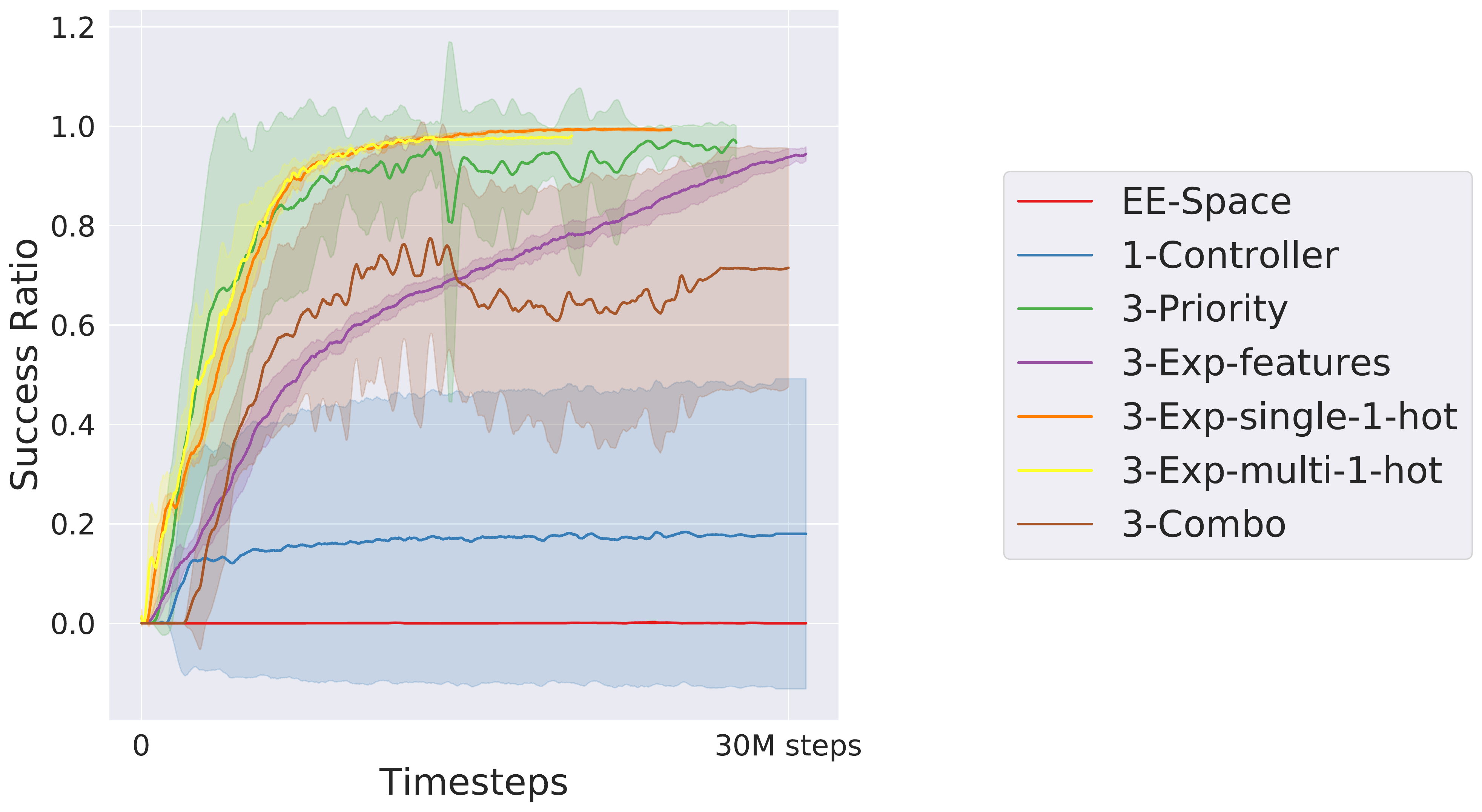}  
  \caption{Franka Hex-Screw Task}
  \label{fig:franka_hex_train_result_appendix}
\end{subfigure}
\begin{subfigure}{.45\textwidth}
  \centering
  \includegraphics[width=.8\textwidth]{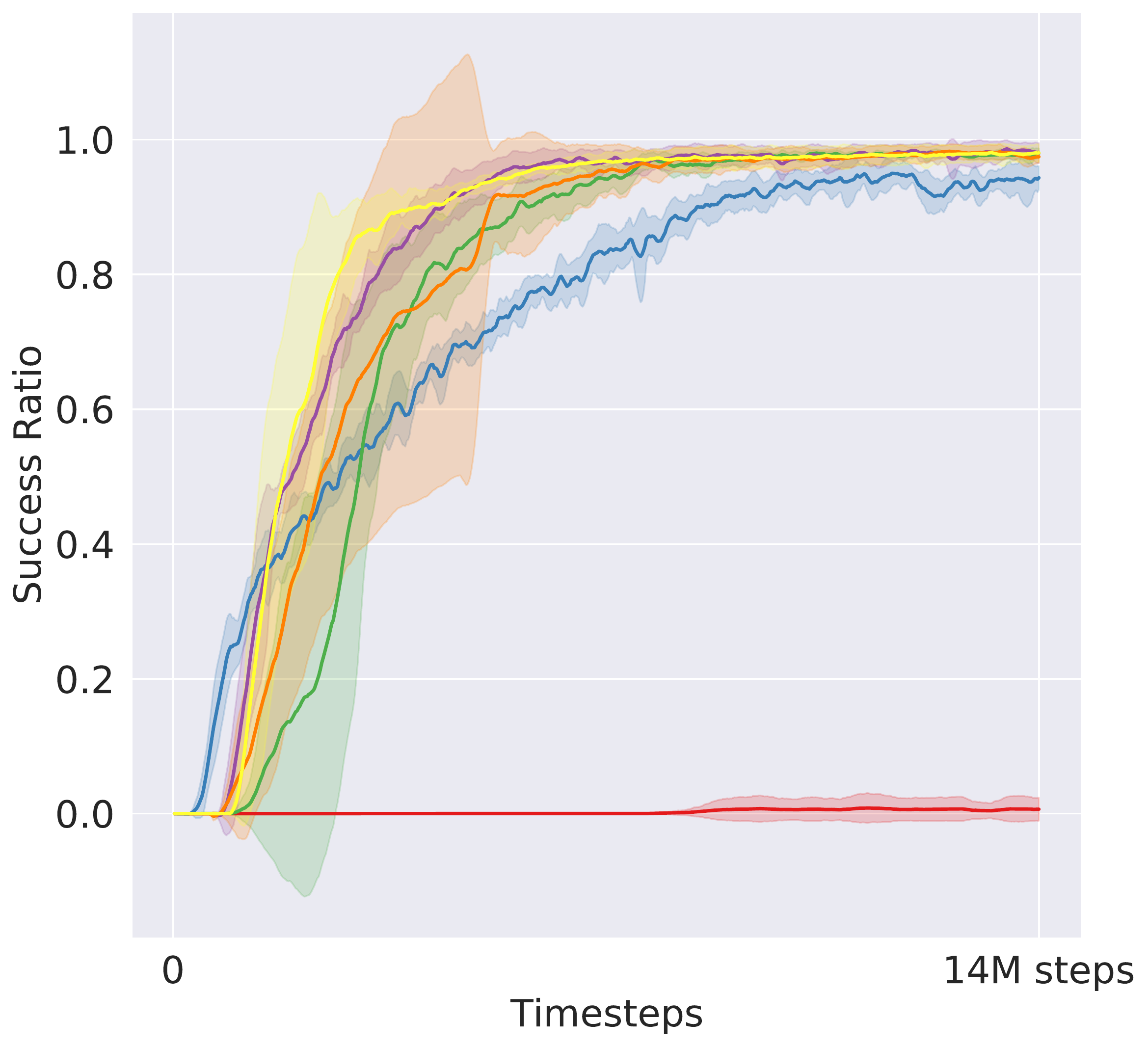}  
  \caption{Franka Door-Opening Task}
  \label{fig:franka_door_train_result_appendix}
\end{subfigure}
\caption{Franka tasks success ratios on training environment configurations during training.}
\label{fig:franka_door_hex_train_results}
\end{figure}

\begin{table}[ht]
\centering
\resizebox{\textwidth}{!}{%
\begin{tabular}{@{}llllllll@{}}
\toprule
Config & EE-Space & 1-Ctrl & 3-Priority & 3-Combo & 3-Exp-Feat & 3-Exp-Single & 3-Exp-Multi \\ \midrule
0.7 & \small{0.0} \scriptsize{(0.00)} & \small{0.05}\scriptsize{(0.14)} & \small{0.69} \scriptsize{(0.44)} & \small{0.45} \scriptsize{(0.43)} & \small{0.95} \scriptsize{(0.04)} & \textbf{\small{0.97} \scriptsize{(0.02)}} & \small{0.96} \scriptsize{(0.035)}  \\
0.8 & \small{0.0} \scriptsize{(0.00)} & \small{0.11}\scriptsize{(0.17)} & \small{0.66} \scriptsize{(0.49)}   & \small{0.43} \scriptsize{(0.43)} &  \small{0.97} \scriptsize{(0.05)} & \small{0.96} \scriptsize{(0.03)} & \textbf{\small{0.98} \scriptsize{(0.01)}} \\
1.1 & \small{0.0} \scriptsize{(0.00)} & \small{0.21}\scriptsize{(0.37)} & \small{0.63} \scriptsize{(0.49)}   & \small{0.43} \scriptsize{(0.36)} &  \small{0.96} \scriptsize{(0.03)} & \textbf{\small{0.98} \scriptsize{(0.03)}} & \small{0.97} \scriptsize{(0.03)} \\
1.2 & \small{0.0} \scriptsize{(0.00)} & \small{0.07} \scriptsize{(0.15)} & \small{0.57} \scriptsize{(0.42)} & \small{0.44} \scriptsize{(0.42)}   &  \textbf{\small{0.97} \scriptsize{(0.04)}}  & \small{0.95}\scriptsize{(0.03)} & \small{0.95} \scriptsize{(0.04)}  \\
1.4 & \small{0.0} \scriptsize{(0.00)} & \small{0.0} \scriptsize{(0.00)} & \small{0.67} \scriptsize{(0.48)}  & \small{0.34} \scriptsize{(0.36)}   & \small{0.92} \scriptsize{(0.03)}  & \small{0.94} \scriptsize{(0.04)} & \textbf{\small{0.95} \scriptsize{(0.03)}} \\
1.5 & \small{0.0} \scriptsize{(0.00)} & \small{0.03} \scriptsize{(0.14)} & \small{0.54} \scriptsize{(0.46)}  & \small{0.24} \scriptsize{(0.36)}  & \textbf{\small{0.92} \scriptsize{(0.04)}}  & \small{0.90} \scriptsize{(0.05)} & \textbf{\small{0.92} \scriptsize{(0.03)}}  \\
\bottomrule \\
\end{tabular}%
}
    \caption{Franka Hex-Screw mean success across all test environment configurations. Parentheses denote standard deviation across 8 seeds.}
    \label{tab:franka_hex_screw_result}
\end{table}

Figure~\ref{fig:franka_hex_train_result_appendix} plots the mean success rates for all the different approaches (including using controller features) during training. 
Since performance on all three train configurations (wrench and screw sizes) is very similar, we report one plot which averages the result for all the configurations.
As seen in Figure~\ref{fig:franka_hex_train_result_appendix}, EE-Space is not able to learn the task. 
While EE-Space policies can bring the wrench close to the screw, it does not achieve proper alignment and insertion, nor does it apply sufficient downard force, all of which are necessary to accomplish the task.
Similarly, 1-Ctrlr also performs poorly. 
This is expected, since the task requires the use of multiple controllers \emph{i.e.} force or position controller into the screw object while also rotating the wrench simultaneously.
For approaches that use multiple object-axis controllers together, we find that the expand-MDP approaches perform the best, robustly learning the task each time. 
All the other approaches suffer from large variance in task performance.

Table~\ref{tab:franka_hex_screw_result} visualizes the result for each of the $6$ different test configurations.  
Each test configuration uses a different wrench and screw scale.
Our proposed approach is able to generalize to the different test configurations, achieving $\ge 0.9$ success rate for all configurations. 
Although 3-Priority performs well in training, its test performance is slightly poorer.
This is because some of the learned policies (seeds) fail to generalize well to any of the test configurations, while the remaining seeds perform as well as our Expanded-MDP approaches. 
This variance in performance of the learned policies leads to lower mean success rate for 3-Priority.
1-Ctrlr fails to work well on any of the test configurations, which is expected given its poor training performance.

\subsection{Franka Door-Opening}


Figure~\ref{fig:franka_door_train_result_appendix} shows the average success rate in all train environments for the Door-Open. 
All methods except EE-Space are able to learn this task. 
One reason for this is that object-centric controllers make exploration in this task much more efficient than directly using the end-effector space. 
Although the EE-Space policy is able to grasp the handle, it fails to turn and pull.
Table~\ref{tab:door_open_gen_result} shows quantitative results on test environments.
Methods that use 3 controllers have very similar performance and perform better than 1-Ctrlr.
Using multiple controllers is beneficial for this task.
When turning the handle, the robot needs to learn to rotate the gripper and press down at the same time; when opening the door, the robot needs to simultaneously press the handle and pull it open. 
Since the reward function contains separate rewards for approaching the handle, turning the handle, and opening the door, the performance differences are due to the complexity of the task and not a lack of informative reward signals.

\begin{table}[ht]
\centering
\resizebox{\textwidth}{!}{%
\begin{tabular}{@{}llllllll@{}}
\toprule
Config & EE-Space & 1-Ctrl & 3-Priority & 3-Combo & 3-Exp-Feat & 3-Exp-Single & 3-Exp-Multi \\ \midrule
A & \small{0.13} \scriptsize{(0.13)} & \small{0.87}\scriptsize{(0.06)} & \small{0.93} \scriptsize{(0.08)} & \small{0.97} \scriptsize{(0.01)} & \textbf{\small{0.99} \scriptsize{(0.01)}} & \textbf{\small{0.99} \scriptsize{(0.01)}} & \textbf{\small{0.99} \scriptsize{(0.01)}}  \\
B & \small{0.00} \scriptsize{(0.00)} & \small{0.96} \scriptsize{(0.02)} & \textbf{\small{0.99} \scriptsize{(0.01)}} & \textbf{\small{0.99} \scriptsize{(0.01)}} & \textbf{\small{0.99} \scriptsize{(0.01)}}  & \textbf{\small{0.99}\scriptsize{(0.01)}} & \textbf{\small{0.99} \scriptsize{(0.02)}}  \\
C & \small{0.00} \scriptsize{(0.00)} & \small{0.93} \scriptsize{(0.03)} & \small{0.99} \scriptsize{(0.01)} & \small{0.99} \scriptsize{(0.01)} &  \textbf{\small{1.00} \scriptsize{(0.00)}}  & \small{0.99} \scriptsize{(0.01)} & \small{0.99} \scriptsize{(0.01)}  \\
\bottomrule \\
\end{tabular}%
}
\caption{Franka Door-Opening mean success on test environment configurations. Parentheses denote standard deviation across 8 seeds.}
\label{tab:door_open_gen_result}
\end{table}


\section{Controller Selection Analyses}

We perform an ablation study to better understand the effects of algorithmic choices in our proposed approach. First, we analyze the effects of controller selection frequency, \emph{i.e.}, we analyze the effect of $T$, where $T$ is the number of steps for which object-axes controllers are run before the RL policy is queried again.
Second, we qualitatively evaluate the learned controller selection policy by visualizing the learned policies. 
For both of these settings we use the Block Fit task.

\begin{figure}
    \centering
    \includegraphics[width=0.8\linewidth]{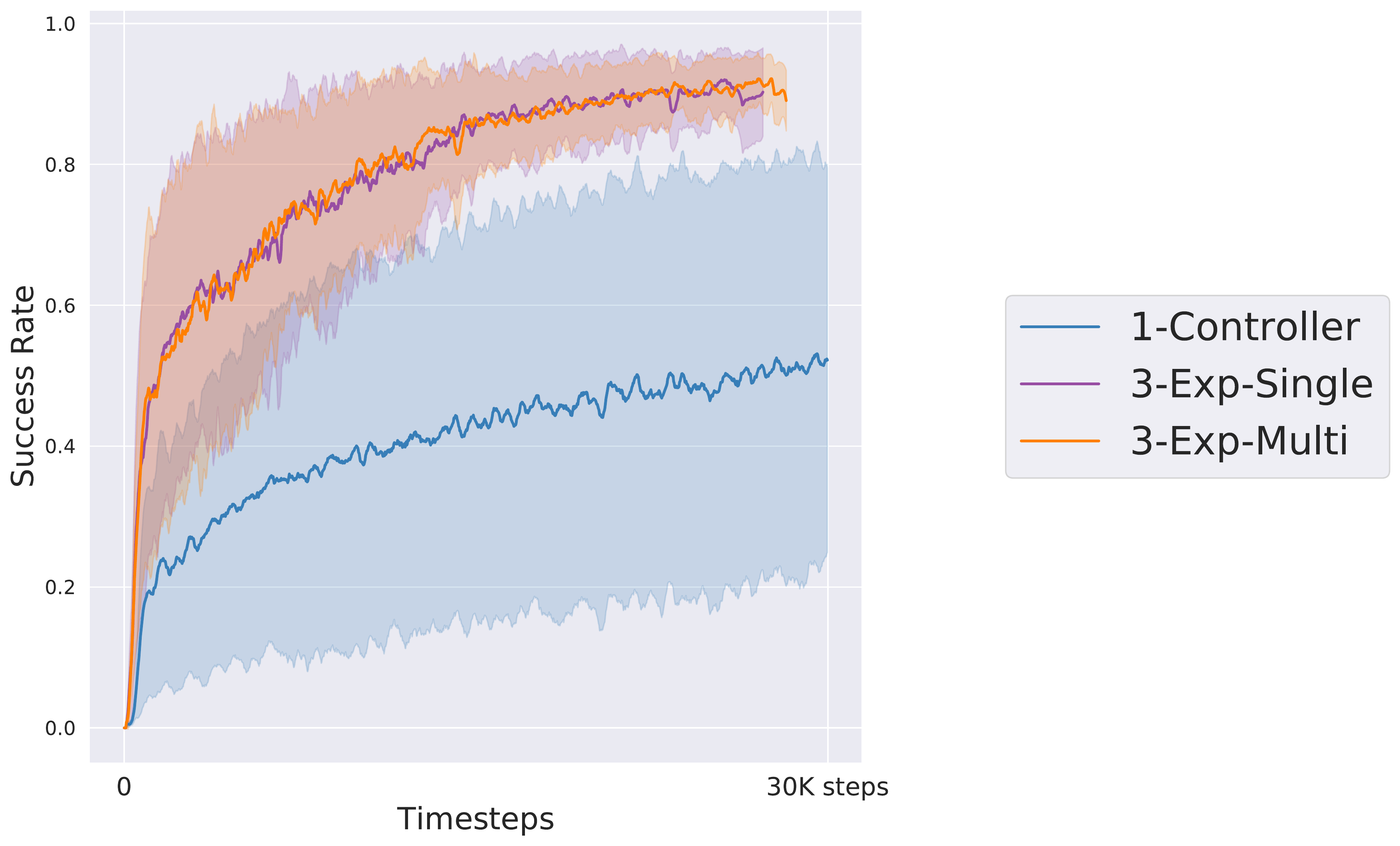}
    \caption{\textbf{Controller Selection Frequency:} Success rate for Block Fit task when object axes-controllers are run for $T=80$ steps. Results averaged over 4 seeds (instead of usual 8).}
    \label{fig:ctrlr_selection_frequency}
\end{figure}

\subsection{Controller Selection Frequency}
We evaluate how  the controller selection frequency $T$ affects the learning performance. For all previous experiments we use $T = 10$, \emph{i.e.}, the object-axes controllers are run only for a few (10) steps. 
Although switching controllers frequently allows the RL policy to be more expressive, this comes at the associated cost of higher sample complexity.
In this experiment we evaluate learning performance when controllers are allowed to run for much larger steps \emph{i.e.} $T=80$. 
To keep the overall simulation time fixed, we simultaneously reduce the maximum number of steps the RL policy is run, \emph{i.e.} we reduce the episode length of the MDP $T_{\text{MDP}}$.
This is important since running both the controllers and the RL policy for large number of steps is computationally prohibitive, since the total number of steps taken in the simulator is $T \times T_{\text{MDP}}$.  We set $T_{\text{MDP}} = 15$ for this experiment.

Figure~\ref{fig:ctrlr_selection_frequency} plots the average success rate for all train configurations on the Block Fit task with $T = 80$. As seen above, our proposed expand-MDP based approaches are able to perform quite well. 
Alternately, selecting only one-controller (1-Controller) at each time step performs poorly as compared to $T=10$ (Figure~\ref{fig:block2d_fit_train_results}). 
This shows the advantage of being able to use multiple object-axes controllers in parallel. With small $T$ the 1-Controller policy is able to complete tasks by quickly switching between different controllers. Since this is not possible with a larger $T$ value, its performance decreases.
This emphasizes the importance of using multiple-controllers in parallel.
Additionally, Figure~\ref{fig:ctrlr_selection_frequency} shows that the Expanded-MDP based approaches are able to learn to perform the task in $30K$ steps only. This is significantly better than the $\sim 10M$ steps required for $T=10$ (Figure~\ref{fig:block2d_fit_train_results}).


\subsection{Controller Selection Visualization}

\begin{figure}[!ht]
\centering
\begin{subfigure}[b]{0.95\textwidth}
    \centering
    \includegraphics[width=0.9\linewidth]{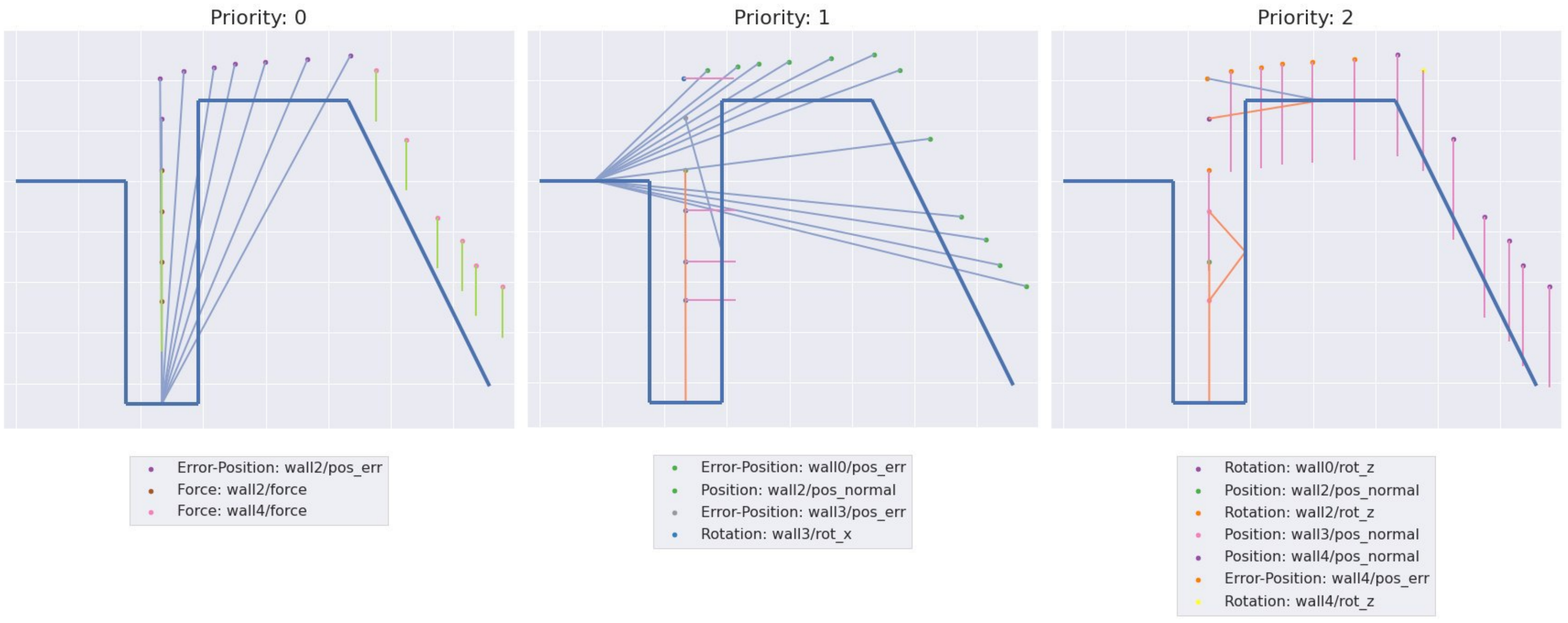}
    \caption*{}
    \label{fig:qualitative_fit_90_diff_8}
\end{subfigure}

\begin{subfigure}[b]{0.95\textwidth}
    \centering
    \includegraphics[width=0.9\linewidth]{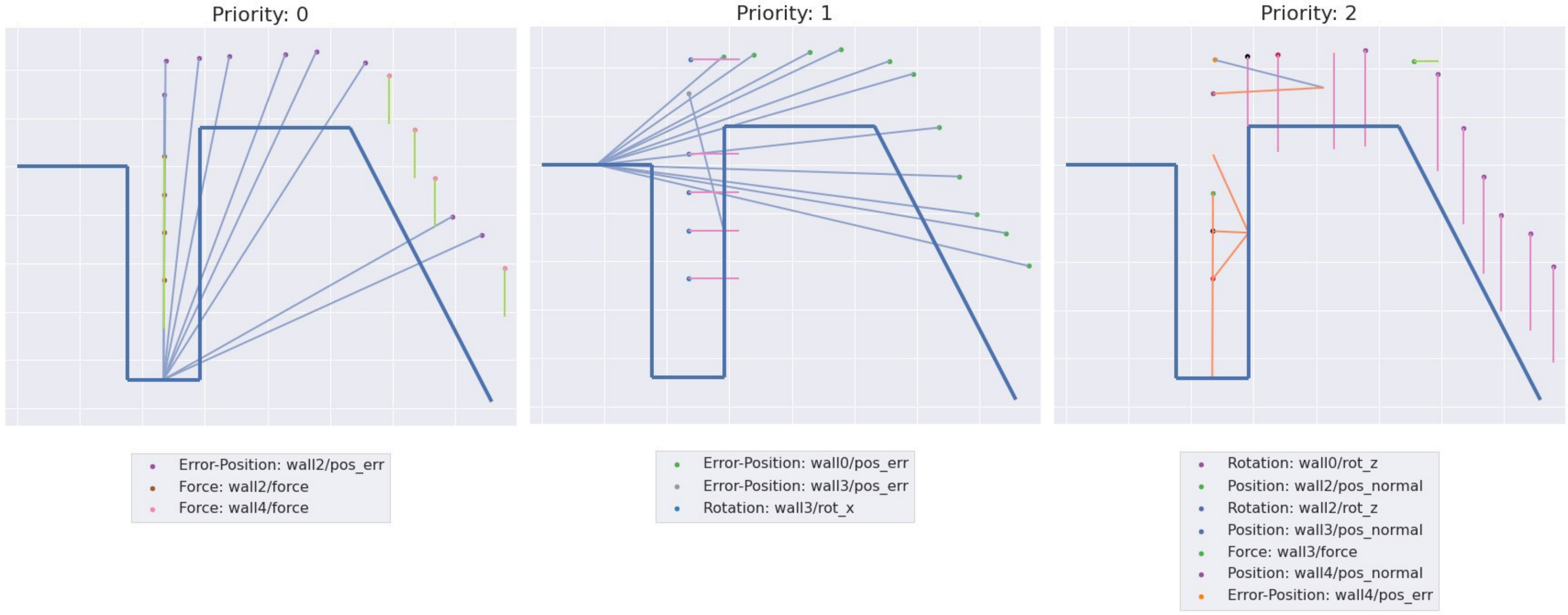}
    \caption*{}
    \label{fig:qualitative_fit_90_diff_4} 
\end{subfigure}
\caption{\textbf{Controller Selection Visualization} for Block Fit during Task Execution.
The thick blue lines show the different walls in the environment. The dots represent the block position at each step. While the arrows represent the wall object used by the selected controller.
The left most plot shows the top priority (priority: 0) controller being selected, while the right most plot shows the controllers with lowest priority (priority: 2). Top and bottom rows are two different train configurations (A and B from Figure~\ref{fig:block2d_test_all}).
}
\label{fig:qualitative_results_block_fit}

\end{figure}

Figure~\ref{fig:qualitative_results_block_fit} plots the controllers the policy selects along the block trajectory for two different train configuration of Block Fit.
For the highest priority controller, the policy tends to select the one that attracts the block toward the target wall.
Interestingly, the second priority controllers are associated with a different wall, i.e the left most wall.
This shows that the RL policy learns to combine controllers across different objects (walls).
For the initial part of the trajectory, the RL policy learns to rotate (priority 0) and move (priority 1) the block simultaneously. 
This composition of different behaviors is important for the policy to accomplish the task as fast as possible. 
In addition, the policy chooses from a few set of controllers for both priority 0 and priority 1, while it chooses from a large set of controllers for priority 2. 
This is because many different choices for the priority 2 controller would often have little to no effect, \emph{e.g.} if both priority 0 and 1 controllers are position or force controllers, then choosing an additional position or force controller for priority 2 will likely have no effect. 
Thus, it is hard for the policy to learn the appropriate effect for lower priority controllers.

\end{appendices}